%% file: main.tex
\documentclass[10pt,twocolumn,letterpaper]{article}

\usepackage{iccv}              % To produce the CAMERA-READY version

\usepackage{epsfig}
\usepackage{graphicx}
\usepackage{amsmath}
\usepackage{amssymb}

\usepackage{booktabs}
\usepackage{multirow}

\usepackage{colortbl}
\usepackage{xcolor}
\usepackage{multirow}
\usepackage{booktabs}
\usepackage{pifont}
\usepackage{bbding}
\usepackage{tcolorbox}


\definecolor{iccvblue}{rgb}{0.21,0.49,0.74}
\usepackage[pagebackref,breaklinks,colorlinks,allcolors=iccvblue]{hyperref}

\def\paperID{2582} % *** Enter the Paper ID here
\def\confName{ICCV}
\def\confYear{2025}

\title{Uncertainty-o: One Model-agnostic Framework\\ for Unveiling Uncertainty in Large Multimodal Models}

\author{
    Ruiyang Zhang ${^{1}}$ \quad Hu Zhang${^{2}}$ \quad
    Hao Fei${^{3}}$
    \quad
    Zhedong Zheng${^1}$\thanks{Correspondence to zhedongzheng@um.edu.mo.}
    \\
     $^1$ FST and ICI, University of Macau, China \\ $^2$ CSIRO Data61, Australia \quad $^3$ National University of Singapore
}

\begin{document}
\maketitle
% \input{sec/0_abstract}    
% \input{sec/1_intro}
% \input{sec/2_related}
% \input{sec/3_method}
% \input{sec/4_exp}
% \input{sec/5_conclusion}

%%%%%%%%% ABSTRACT
\begin{abstract}

Large Multimodal Models (LMMs), harnessing the complementarity among diverse modalities, are often considered more robust than pure Language Large Models (LLMs); yet do LMMs know what they do not know? There are three key open questions remaining: (1) how to evaluate the uncertainty of diverse LMMs in a unified manner, (2) how to prompt LMMs to show its uncertainty, and (3) how to quantify uncertainty for downstream tasks.

%whether LMMs provide consistent answers when the same question is presented in different formats

In an attempt to address these challenges, we introduce \textbf{Uncertainty-o}: (1) a model-agnostic framework designed to reveal uncertainty in LMMs regardless of their modalities, architectures, or capabilities, (2) an empirical exploration of multimodal prompt perturbations to uncover LMM uncertainty, offering insights and findings, and (3) derive the formulation of multimodal semantic uncertainty, which enables quantifying uncertainty from multimodal responses. Experiments across 18 benchmarks spanning various modalities and 10 LMMs (both open- and closed-source) demonstrate the effectiveness of Uncertainty-o in reliably estimating LMM uncertainty, thereby enhancing downstream tasks such as hallucination detection, hallucination mitigation, and uncertainty-aware Chain-of-Thought reasoning.

\end{abstract}

%%%%%%%%% BODY TEXT
\section{Introduction}

Large Multi-Modal Models (LMMs)~\cite{caffagni2024revolution,carolan2024review} significantly expand the boundaries of human-made intelligence. \textbf{(1) Rich modalities are continuously integrated.} Beyond the initial success of the `Text-in-Text-out' paradigm~\cite{achiam2023gpt}, LMMs now incorporate images~\cite{liu2023visual,rombach2022high}, videos~\cite{lin2023video,luo2023videofusion}, audio~\cite{radford2023robust,kreuk2022audiogen}, point clouds~\cite{xu2024pointllm,nichol2022point}, IMU~\cite{girdhar2023imagebind}, fMRI~\cite{han2024onellm}, and more~\cite{wu2024next}. \textbf{(2) Diverse model architectures are rapidly evolving.} The foundational designs of Large Language Models (LLMs)~\cite{achiam2023gpt,brown2020language} and Diffusion Models (DMs)~\cite{rombach2022high,ho2020denoising} have been extensively adapted~\cite{liu2023visual} and integrated~\cite{wu2024self} to create more advanced and complex structures. \textbf{(3) Distinct model capabilities have significantly improved.} LMMs can now comprehend intricate multimodal contexts and instructions~\cite{lyu2024unibind,han2024onellm}, while simultaneously generating diverse and high-fidelity content~\cite{wu2024next,zhan2024anygpt}.

\begin{figure}
    \centering
    \includegraphics[width=\linewidth]{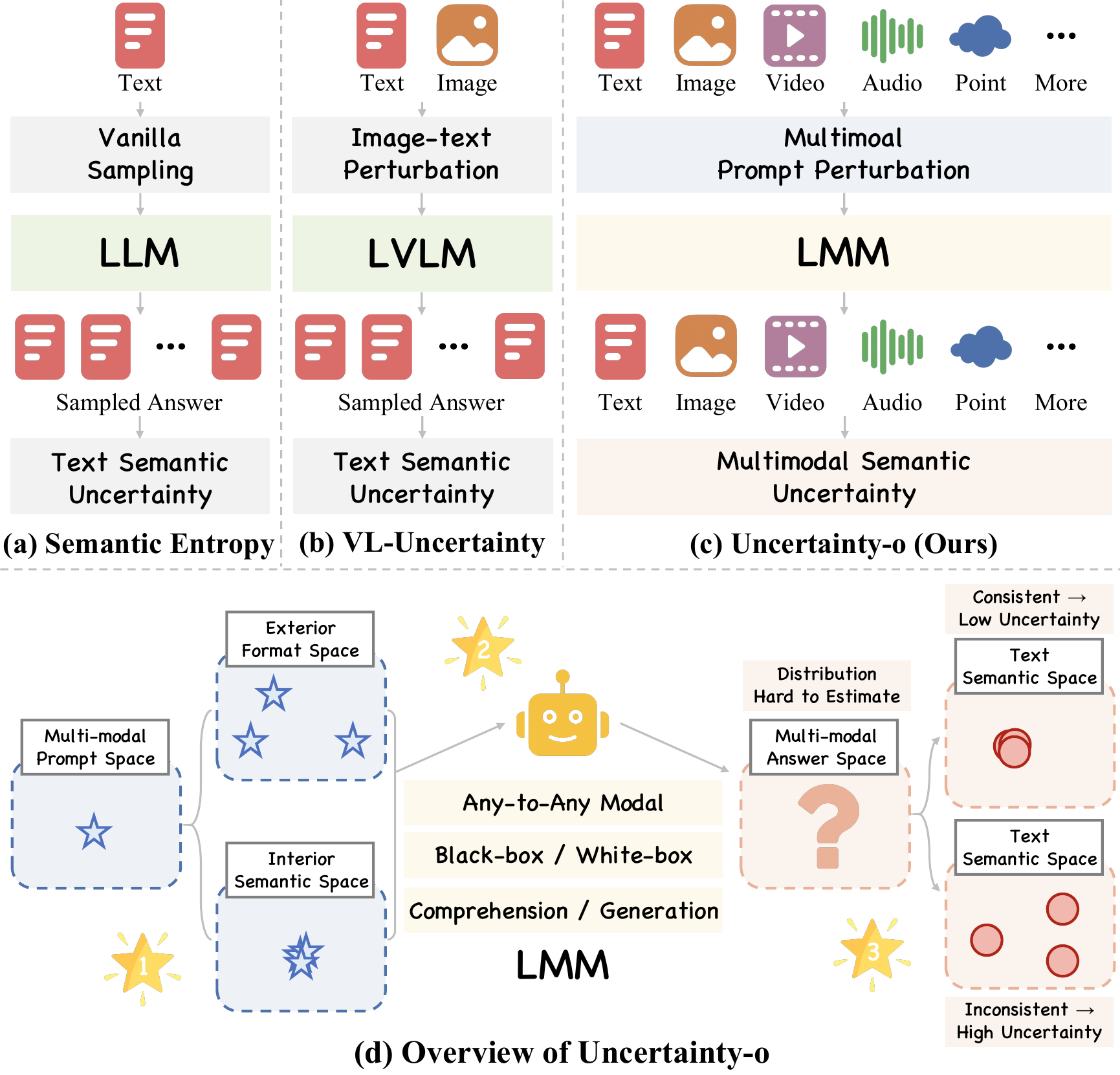}
    \vspace{-20pt}
    \caption{
    \textbf{Comparison with Previous Works and Overview of Uncertainty-o.}
    (a,b) Existing studies focus on single modality or two modalities~\cite{zhang2024vl}; (c) In contrast, our Uncertainty-o captures uncertainty in \textit{large multimodal models} in a model-agnostic manner. It achieves reliable uncertainty estimation via \textit{multimodal prompt perturbation}. 
    (d) In particular, we harnessing \textit{multimodal semantic uncertainty}, which maps multimodal answer space into a unified text semantic space. We then estimate the uncertainty according to the text consistency.
    }
    \label{fig:comparison}
\end{figure}

\begin{figure*}[!ht]
    \vspace{-30pt}
    \centering
    \includegraphics[width=\linewidth]{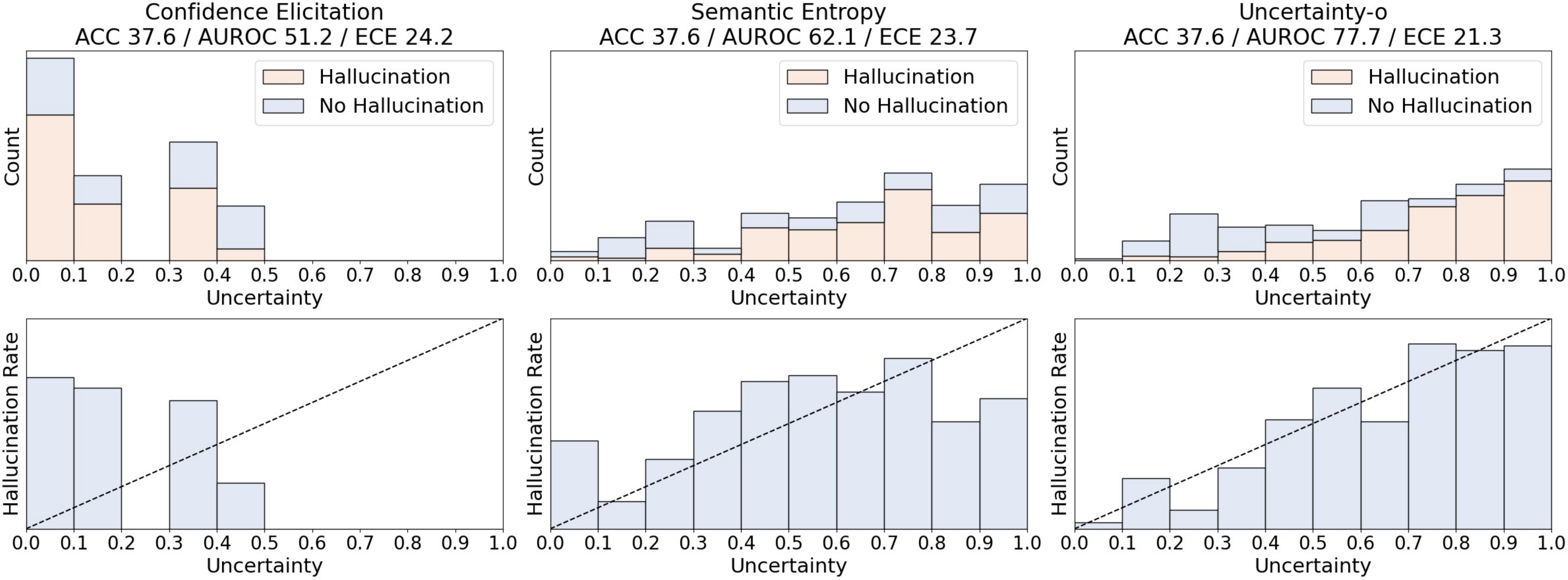}
    \vspace{-15pt}
    \caption{
    \textbf{Reliability of Our Estimated Uncertainty.}
    From comparison with previous methods, we observe that: (1) Compared to Confidence Elicitation~\cite{xiong2023can}, Uncertainty-o effectively avoids the over-confidence problem, which inaccurately assigns low uncertainty (or high confidence) to hallucinatory responses. (2) Compared to Semantic Entropy~\cite{farquhar2024detecting}, Uncertainty-o provides more reliable uncertainty estimation, \eg, the uncertainty is more closely aligned with the error rate in each bin. Results from OneLLM~\cite{han2024onellm} on ClothoV2~\cite{drossos2020clotho}.
    }
    \label{fig:calibration}
    \vspace{-10pt}
\end{figure*}

However, large models~\cite{liu2023visual,achiam2023gpt} still follow a vanilla `Prompt-In-Response-Out' pipeline, which remains raw and primitive compared to human thinking process~\cite{amayuelas2023knowledge,pi2024mllm}. When asked a question or given a task, humans not only provide an answer or take action, but also experience internal feelings: confidence or uncertainty, ease or difficulty~\cite{peterson2019human,collins2023human}. These feelings guide us in predicting outcomes and adjusting our plans accordingly~\cite{callaway2022rational,auletta2023predicting}.  To bridge this gap and explore the psychology of LMMs~\cite{ke2024exploring,griffin2023large}, we focus on one critical property: LMM uncertainty~\cite{aichberger2024many,shorinwa2024survey}.

% In this paper, we aim to answer 3 key questions: \textbf{(1) Can we unveil LMM uncertainty in a model-agnostic manner?} Tailoring specific design for different LMMs can lead to a complex and inflexible framework, limiting adaptability to revolving LMMs~\cite{caffagni2024revolution,carolan2024review}. \textbf{(2) How to reveal the aleatoric uncertainty in multimodal prompts?} Notably, ambiguity and observational noise in prompt can significantly lead to confusion of LMM. Moreover, interaction and relationship between various introduced modalities exacerbate this phenomenon~\cite{lu2023theory,fu2022complexity}. \textbf{(3) How to mine uncertainty from multimodal responses where LMM manifests its epistemic uncertainty?} The cues are hidden in these multimodal responses~\cite{rombach2022high,kreuk2022audiogen,nichol2022point,wu2024next}, \eg, generated images~\cite{rombach2022high} or point clouds~\cite{lee2024rgb2point}, requiring effective methods to extract and quantify LMM epistemic uncertainty.

Despite the strong performance of LMMs, three key open questions remains: \textbf{(1) How to evaluate the uncertainty of diverse LMMs in a unified manner?} 
It is challenging in keeping the same evaluation standard for different LMMs. 
The vanilla solution is to design specific solutions according to the input modality, but can result in a complex and inflexible framework, limiting adaptability to involving new modalities~\cite{caffagni2024revolution,carolan2024review}. 
\textbf{(2) How to prompt LMM show its uncertainty?}  
The naive method is ask the LMM uncertainty via prompt inputs, but it usually fails. It is due to that LMMs, including LLMs, are usually over-confident about their response.
%In contrast, if we ask the LLMs twice about the same question, LMMs sometimes give a different answer. 
%Could we le
%Is LMMs robust against the same prompt in different formats?} 
%The LMM performance is impacted 
%when prompts are composed of multimodal information, it is intriguing to investigate the effect of changing one modality while keeping the others fixed~\cite{lu2023theory,fu2022complexity}. 
\textbf{(3) How can we quantify uncertainty for downstream tasks?} The cues are embedded in multimodal responses~\cite{rombach2022high,kreuk2022audiogen,nichol2022point,wu2024next}, \eg, generated images~\cite{rombach2022high} or point clouds~\cite{lee2024rgb2point}, requiring effective techniques to extract and quantify LMM uncertainty to facilitate downstream tasks.

In response to those substantial questions, we propose Uncertainty-o (see Fig.~\ref{fig:comparison}): \textbf{(1) Uncertainty-o, in a model-agnostic manner, unveils uncertainty in LMMs, regardless of the modalities involved, inherent architecture, and capability focus.} Notably, Uncertainty-o is also scalable and can incorporate emerged modalities, architectures, and capabilities. \textbf{(2) We empirically explore multimodal prompt perturbation for revealing prompt aleatoric uncertainty.} By causing variance at the prompt end, we can observe fluctuations in LMM responses. Higher degree of fluctuation indicates higher level of aleatoric uncertainty, illustrating challenge and complexity of prompt. We provide a comprehensive empirical analysis of perturbation for 5 commonly used modalities, including text, image, video, audio, and point cloud. Finally, we offer our findings and insights, which are modality-agnostic and can serve as a cookbook when designing perturbation for other potential modalities. \textbf{(3) We propose multimodal semantic uncertainty, which can mine LMM inherent epistemic uncertainty from multimodal responses.} Through mapping multimodal responses to text semantic space, the observation of answer semantic distribution becomes addressable. The answer distribution entropy in the text semantic space serve as an effective indicator of epistemic uncertainty. Furthermore, uncertainties of each modality are merged to reveal overall LMM uncertainty.

Through extensive experiments with 18 multimodal benchmarks (including 5 modalities) and 10 LMMs (both open- and closed-source), we validate the effectiveness of Uncertainty-o in accurately capturing the uncertainty of LMMs. The reliable uncertainty estimation (see Fig.~\ref{fig:calibration}) facilitates various downstream tasks, including hallucination detection, hallucination mitigation, and uncertainty-aware CoT.
% remark
In summary, our contributions are as follows:
\begin{itemize}
\item \textbf{Unified Framework to Unveil LMM Uncertainty.} We propose Uncertainty-o, which, in a model-agnostic manner, reveals one critical property in LMM psychology: uncertainty. Uncertainty-o can seamlessly handle LMMs with various modalities, complex model architectures, and distinct capacity emphases.
\item \textbf{Cookbook for Multimodal Prompt Perturbation.} We propose multimodal prompt perturbation and empirically explore how to perturb the prompt to maximize its contribution to revealing prompt \textit{aleatoric uncertainty}. 
\item \textbf{Multimodal Uncertainty in Semantic Space.} We propose multimodal semantic uncertainty, explicitly enabling the mining of inherent \textit{epistemic uncertainty} from multimodal generated content via semantic space mapping.
\end{itemize}

\begin{figure*}
    \centering
    \includegraphics[width=0.95\linewidth]{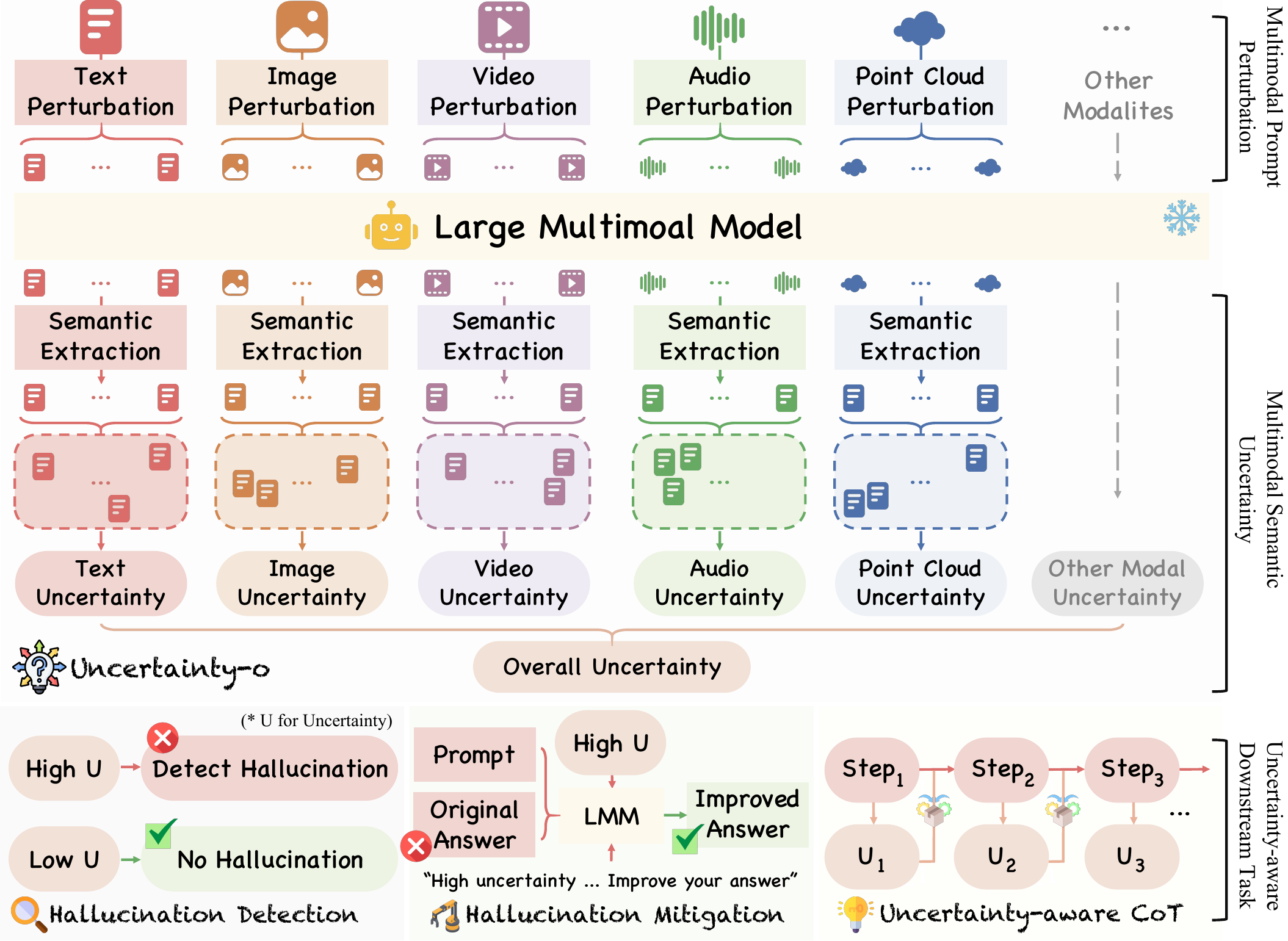}
    \caption{
    \textbf{Pipeline of Our Uncertainty-o.} Given a multimodal prompt and large multimodal models, we perform multimodal prompt perturbation to generate diverse responses. Due to the inherent epistemic uncertainty of these models under perturbation, varied responses are typically obtained. To quantify this uncertainty, we apply semantic clustering on the collected responses and compute their entropy. Specifically, responses are grouped into semantically similar clusters, and the entropy across these clusters is calculated as the final uncertainty measure. Higher entropy indicates greater variability in responses, suggesting lower confidence, while lower entropy reflects higher consistency and thus higher confidence. Finally, we could leverage the uncertainty for hallucination detection, hallucination mitigation and facilitate the chain of thoughts (CoT).
    }
    \label{fig:method}
\end{figure*}

\section{Uncertainty-o}

\subsection{Multimodal Prompt Perturbation}

\noindent \textbf{Definition 3.1 (Response Semantic Distance).} \textit{The LMM $M$ generate reponses for prompts $x_{i}$ and $x_{j}$ (perturbed from original prompt $x$). For perturbed prompts, the predictions from LMM is $y_{i} = M(x_{i})$, and $y_{j} = M(x_{j})$, respectively. We focus on the LMM's \underline{Epistemic Uncertainty}, which is the uncertainty in the model parameters, $\text{Var}(\theta|x)$. Define the prediction difference $D(x)$ as:}
\begin{equation}
D(x) = \| y_{i} - y_{j} \|,
\end{equation}
\textit{where $\|\cdot\|$ denotes $\ell_2$ norm}.

\noindent \textbf{Proposition 3.2.} \textit{Response semantic distance $D(x)$ from prompt perturbation is proportional to the square root of the LMM uncertainty:}
\begin{equation}
D(x) \propto \sqrt{\text{Var}(\theta|x)}.
\end{equation}
\textit{Therefore, the prediction semantic distance $D(x)$ can serve as a measure of the LMM uncertainty under given context.}

\vspace{1mm}

Based on \textit{Proposition 3.2}, we propose semantic-equivalent perturbation for multimodal prompts to induce variations in LMM responses, enabling the capture of LMM uncertainty (see Fig.~\ref{fig:method}). This approach generates semantic-equivalent variations of original prompts across multiple modalities and feeds them into the LMM. The variance in the sampled responses could effectively illustrate LMM uncertainty.
Specifically, we propose multimodal prompt perturbation across five modalities, including text, image, audio, video, and point cloud. Notably, we adopt a progressive perturbation strategy, where each prompt is gradually perturbed from low to high degrees, introducing increasing levels of challenge to the LMM. Perturbations across different modalities are applied simultaneously to further enhance uncertainty estimation.

% we also adopt a progressive approach to multimodal prompt perturbation. Each prompt is perturbed gradually, from low to high degrees of perturbation. This strategy introduces different levels of challenge to the LMM. Additionally, perturbations of various input modalities are conducted simultaneously.

In practice, different perturbation techniques are applied depending on the modality. For text prompts, both LLM-based rephrasing and rule-based methods, such as word swapping, are used. For image prompts, spatial transformations like rotation and attribute distortions such as blurring or brightness adjustments are applied. In the case of audio prompts, key elements like volume, pitch, and timbre are perturbed, with additional temporal adjustments such as temporal shifting. For video prompts, both spatial and temporal perturbations are applied. Temporal perturbations include techniques like temporal cropping and frame dropping, while spatial perturbations involve applying image perturbation techniques to each video frame. For point cloud prompts, 3D characteristics such as point density and shape are perturbed, using techniques like random sampling and point cloud jittering.

\begin{table*}[!ht]
    \centering
    \fontsize{8}{9}\selectfont
    \setlength{\tabcolsep}{2.4mm}
    % \resizebox{0.9\linewidth}{!}{
    \begin{tabular}{l|ccc|ccc|ccc}
        \toprule
        \multicolumn{10}{c}{\textit{Image Comprehension}} \\
        \midrule
        \multirow{2}{*}{Method} & \multicolumn{3}{c|}{LLaVABench~\cite{liu2023visual}} & \multicolumn{3}{c|}{MMVet~\cite{yu2023mm}} & \multicolumn{3}{c}{CoCoCap~\cite{chen2015microsoft}} \\ 
        ~ & AUROC$\uparrow$ & AURAC$\uparrow$ & ECE$\downarrow$ & AUROC$\uparrow$ & AURAC$\uparrow$ & ECE$\downarrow$ & AUROC$\uparrow$ & AURAC$\uparrow$ & ECE$\downarrow$ \\ 
        \midrule
        Confidence Elicitation~\cite{xiong2023can} & 54.0 & 78.1 & \underline{11.3} & 52.3 & 55.8 & 13.1 & 51.2 & 71.0 & 23.1 \\ 
        GAVIE~\cite{liu2023mitigating} & 45.3 & 75.2 & 23.8 & 55.4 & 45.2 & 9.4 & 57.7 & \underline{79.6} & 28.9 \\ 
        Semantic Entropy~\cite{farquhar2024detecting} & \underline{65.8} & \underline{87.4} & 14.9 & \underline{60.6} & \underline{56.9} & \underline{8.8} & \underline{60.1} & 78.4 & \underline{6.9} \\ 
        \rowcolor{gray!15} \textbf{Uncertainty-o (ours)} & \textbf{68.0} & \textbf{91.3} & \textbf{10.1} & \textbf{63.5} & \textbf{66.4} & \textbf{8.2} & \textbf{65.5} & \textbf{82.5} & \textbf{5.9} \\ 
        \midrule
        \multicolumn{10}{c}{\textit{Video Comprehension}} \\
        \midrule
        \multirow{2}{*}{Method} & \multicolumn{3}{c|}{MSRVTTQA~\cite{xu2017video}} & \multicolumn{3}{c|}{MSVDQA~\cite{xu2017video}} & \multicolumn{3}{c}{NextQA~\cite{xiao2021next}} \\ 
        ~ & AUROC$\uparrow$ & AURAC$\uparrow$ & ECE$\downarrow$ & AUROC$\uparrow$ & AURAC$\uparrow$ & ECE$\downarrow$ & AUROC$\uparrow$ & AURAC$\uparrow$ & ECE$\downarrow$ \\ 
        \midrule
        Confidence Elicitation~\cite{xiong2023can} & \underline{53.5} & 32.6 & \underline{12.2} & \underline{51.0} & \underline{32.8} & \underline{16.1} & 51.5 & \underline{65.9} & \underline{10.9} \\ 
        GAVIE~\cite{liu2023mitigating} & 50.1 & 33.9 & 18.2 & 39.6 & 30.1 & 20.5 & 47.4 & 60.8 & 11.4 \\ 
        Semantic Entropy~\cite{farquhar2024detecting} & 51.6 & \underline{35.1} & 14.7 & 47.1 & 32.7 & 18.6 & \underline{59.1} & 62.4 & 12.6 \\ 
        \rowcolor{gray!15} \textbf{Uncertainty-o (ours)} & \textbf{55.4} & \textbf{39.2} & \textbf{10.8} & \textbf{52.4} & \textbf{36.3} & \textbf{14.5} & \textbf{67.4} & \textbf{69.0} & \textbf{7.4} \\ 
        \midrule
        \multicolumn{10}{c}{\textit{Audio Comprehension}} \\
        \midrule
        \multirow{2}{*}{Method} & \multicolumn{3}{c|}{ClothoV2~\cite{drossos2020clotho}} & \multicolumn{3}{c|}{ClothoAQA~\cite{lipping2022clotho}} & \multicolumn{3}{c}{AudioCaps~\cite{kim2019audiocaps}} \\ 
        ~ & AUROC$\uparrow$ & AURAC$\uparrow$ & ECE$\downarrow$ & AUROC$\uparrow$ & AURAC$\uparrow$ & ECE$\downarrow$ & AUROC$\uparrow$ & AURAC$\uparrow$ & ECE$\downarrow$ \\ 
        \midrule
        Confidence Elicitation~\cite{xiong2023can} & 51.2 & 34.1 & 24.2 & 49.2 & 43.5 & 29.0 & 50.9 & 30.2 & 18.6 \\ 
        GAVIE~\cite{liu2023mitigating} & 51.1 & 31.9 & 30.6 & 53.1 & 46.7 & 23.9 & 47.7 & 34.8 & 19.7 \\ 
        Semantic Entropy~\cite{farquhar2024detecting} & \underline{62.1} & \underline{38.5} & \underline{23.7} & \underline{56.7} & \underline{50.2} & \underline{17.9} & \underline{50.9} & \underline{42.4} & \underline{16.1} \\ 
        \rowcolor{gray!15} \textbf{Uncertainty-o (ours)} & \textbf{77.7} & \textbf{44.1} & \textbf{21.3} & \textbf{68.1} & \textbf{51.0} & \textbf{17.7} & \textbf{56.7} & \textbf{46.6} & \textbf{12.4} \\
        \midrule
        \multicolumn{10}{c}{\textit{Point Cloud Comprehension}} \\
        \midrule
        \multirow{2}{*}{Method} & \multicolumn{3}{c|}{ModelNet~\cite{wu20153d}} & \multicolumn{3}{c|}{ShapeNet~\cite{chang2015shapenet}} & \multicolumn{3}{c}{Objaverse~\cite{deitke2023objaverse}} \\ 
        ~ & AUROC$\uparrow$ & AURAC$\uparrow$ & ECE$\downarrow$ & AUROC$\uparrow$ & AURAC$\uparrow$ & ECE$\downarrow$ & AUROC$\uparrow$ & AURAC$\uparrow$ & ECE$\downarrow$ \\ 
        \midrule
        Confidence Elicitation~\cite{xiong2023can} & 47.1 & 78.0 & 36.1 & 46.4 & 72.1 & 39.7 & 49.5 & 28.0 & \underline{23.1} \\ 
        GAVIE~\cite{liu2023mitigating} & 46.0 & \underline{84.5} & 43.5 & 51.1 & \underline{75.4} & 37.5 & 49.8 & 35.1 & 29.4 \\ 
        Semantic Entropy~\cite{farquhar2024detecting} & \underline{52.7} & 79.2 & \underline{39.8} & \underline{54.6} & 73.6 & \underline{32.6} & \underline{50.1} & \underline{38.5} & 26.0 \\ 
        \rowcolor{gray!15} \textbf{Uncertainty-o (ours)} & \textbf{66.5} & \textbf{92.1} & \textbf{31.9} & \textbf{60.6} & \textbf{78.0} & \textbf{31.7} & \textbf{56.9} & \textbf{43.8} & \textbf{18.7} \\
        \bottomrule
    \end{tabular}
    % }
    \vspace{-5pt}
    
    \caption{
    \textbf{Comprehension Hallucination Detection Results.} 
    Uncertainty-o consistently outperforms strong baselines by a clear margin in LMM comprehension hallucination detection. For LMMs, We utilize InternVL~\cite{chen2024internvl}, VideoLLaMA~\cite{zhang2023video}, OneLLM~\cite{han2024onellm}, PointLLM~\cite{xu2024pointllm} for image, video, audio, point cloud comprehension. The best results are in \textbf{bold}, and the second-best results are \underline{underlined}.
    }
    \label{tab:main_comprehension}
\end{table*}

\begin{table}[!t]
    \vspace{-10pt}
    \centering
    \fontsize{8}{9}\selectfont
    \setlength{\tabcolsep}{2.4mm}
    % \resizebox{0.8\linewidth}{!}{
    \begin{tabular}{l|ccc}
        \toprule
        \multirow{2}{*}{Method} & \multicolumn{3}{c}{MMVet~\cite{yu2023mm} w. GPT4o~\cite{hurst2024gpt}} \\ 
        ~ & AUROC$\uparrow$ & AURAC$\uparrow$ & ECE$\downarrow$ \\ 
        \midrule
        Confidence Elicitation~\cite{xiong2023can} & \underline{49.5} & \underline{67.8} & 11.7 \\
        GAVIE~\cite{liu2023mitigating} & 49.0 & 65.6 & \underline{10.4} \\
        Semantic Entropy~\cite{farquhar2024detecting} & 52.8 & 64.9 & 11.1 \\
        \rowcolor{gray!15} \textbf{Uncertainty-o (ours)} & \textbf{56.1} & \textbf{75.8} & \textbf{8.6} \\
        \midrule
        \midrule
        \multirow{2}{*}{Method} & \multicolumn{3}{c}{MSRVTTQA~\cite{xu2017video} w. QwenVLMax~\cite{bai2023qwenvlversatilevisionlanguagemodel}} \\ 
        ~ & AUROC$\uparrow$ & AURAC$\uparrow$ & ECE$\downarrow$ \\ 
        \midrule
        Confidence Elicitation~\cite{xiong2023can} & 54.7 & 39.9 & \underline{11.5} \\
        GAVIE~\cite{liu2023mitigating} & 55.9 & 40.1 & 15.6  \\
        Semantic Entropy~\cite{farquhar2024detecting} & \underline{56.8} & \underline{41.5} & 12.8 \\
        \rowcolor{gray!15} \textbf{Uncertainty-o (ours)} & \textbf{60.1} & \textbf{47.5} & \textbf{8.9} \\
        \bottomrule
    \end{tabular}
    % }
    \vspace{-5pt}
    
    \caption{
    \textbf{Hallucination Detection for Closed-Source LMMs.} We leverage GPT4o~\cite{hurst2024gpt} and QwenVLMax~\cite{bai2023qwenvlversatilevisionlanguagemodel} for image and video comprehension respectively.
    }
    \label{tab:main_close}
    \vspace{-10pt}
\end{table}

\begin{table}[!t]
    \vspace{-10pt}

    \centering
    \fontsize{8}{9}\selectfont
    \setlength{\tabcolsep}{2.4mm}
    % \resizebox{0.8\linewidth}{!}{
    \begin{tabular}{l|ccc}
        \toprule
        \multirow{2}{*}{Method} & \multicolumn{3}{c}{\textit{Medical Diagnosis}} \\ 
        ~ & AUROC$\uparrow$ & AURAC$\uparrow$ & ECE$\downarrow$ \\ 
        \midrule
        Confidence Elicitation~\cite{xiong2023can} & 42.1 & 30.9 & 24.5 \\
        GAVIE~\cite{liu2023mitigating} & 44.2 & 33.6 & 31.5 \\
        Semantic Entropy~\cite{farquhar2024detecting} & \underline{52.4} & \underline{34.5} & \underline{24.1} \\
        \rowcolor{gray!15} \textbf{Uncertainty-o (ours)} & \textbf{59.0} & \textbf{40.6} & \textbf{15.3} \\
        \midrule
        \midrule
        \multirow{2}{*}{Method} & \multicolumn{3}{c}{\textit{Embodied Robot}} \\ 
        ~ & AUROC$\uparrow$ & AURAC$\uparrow$ & ECE$\downarrow$ \\ 
        \midrule
        Confidence Elicitation~\cite{xiong2023can} & 40.1 & 49.6 & \underline{18.9} \\
        GAVIE~\cite{liu2023mitigating} & 45.2 & 53.6 & 23.5 \\
        Semantic Entropy~\cite{farquhar2024detecting} & \underline{56.2} & \underline{55.9} & 21.1 \\
        \rowcolor{gray!15} \textbf{Uncertainty-o (ours)} & \textbf{58.2} & \textbf{66.5} & \textbf{12.2} \\
        \bottomrule
    \end{tabular}
    % }
    \vspace{-5pt}
    
    \caption{
    \textbf{Hallucination Detection for Safety-Critic Tasks.}
    We harness MIMICCXR~\cite{johnson2019mimic} for medical image diagnosis and OpenEQA~\cite{majumdar2024openeqa} for video embodied QA.
    }
    \label{tab:main_safety}
    \vspace{-10pt}
\end{table}

\begin{table}[!t]
    \centering
    \vspace{-20pt}
    
    \fontsize{8}{9}\selectfont
    \setlength{\tabcolsep}{2.4mm}
    % \resizebox{0.8\linewidth}{!}{
    \begin{tabular}{l|ccc}
        \toprule
        \multicolumn{4}{c}{\textit{Image Generation}} \\
        \midrule
        \multirow{2}{*}{Method} & \multicolumn{3}{c}{Flickr~\cite{plummer2015flickr30k}} \\ 
        ~ & AUROC$\uparrow$ & AURAC$\uparrow$ & ECE$\downarrow$ \\ 
        \midrule
        GAVIE~\cite{liu2023mitigating} & \underline{52.3} & 67.9 & \underline{12.7} \\ 
        Semantic Entropy~\cite{farquhar2024detecting} & 50.8 & \underline{70.1} & 21.5 \\ 
        \rowcolor{gray!15} \textbf{Uncertainty-o (ours)} & \textbf{59.5} & \textbf{74.5} & \textbf{8.3} \\ 
        \midrule
        \multicolumn{4}{c}{\textit{Video Generation}} \\
        \midrule
        \multirow{2}{*}{Method} & \multicolumn{3}{c}{MSRVTT~\cite{xu2016msr}} \\ 
        ~ & AUROC$\uparrow$ & AURAC$\uparrow$ & ECE$\downarrow$ \\ 
        \midrule
        GAVIE~\cite{liu2023mitigating} & 51.6 & 24.1 & \underline{21.7} \\ 
        Semantic Entropy~\cite{farquhar2024detecting} & \underline{54.7} & \underline{30.9} & 23.0 \\ 
        \rowcolor{gray!15} \textbf{Uncertainty-o (ours)} & \textbf{61.1} & \textbf{38.9} & \textbf{15.5} \\
        \midrule
        \multicolumn{4}{c}{\textit{Audio Generation}} \\
        \midrule
        \multirow{2}{*}{Method} & \multicolumn{3}{c}{VCTK~\cite{Yamagishi2019CSTRVC}} \\ 
        ~ & AUROC$\uparrow$ & AURAC$\uparrow$ & ECE$\downarrow$ \\ 
        \midrule
        GAVIE~\cite{liu2023mitigating} & 46.0 & \underline{74.6} & 21.5 \\ 
        Semantic Entropy~\cite{farquhar2024detecting} & \underline{48.1} & 72.8 & \underline{20.6} \\ 
        \rowcolor{gray!15} \textbf{Uncertainty-o (ours)} & \textbf{53.5} & \textbf{81.4} & \textbf{11.3} \\ 
        \midrule
        \multicolumn{4}{c}{\textit{Point Cloud Generation}} \\
        \midrule
        \multirow{2}{*}{Method} & \multicolumn{3}{c}{Pix3D~\cite{sun2018pix3d}} \\ 
        ~ & AUROC$\uparrow$ & AURAC$\uparrow$ & ECE$\downarrow$ \\ 
        \midrule
        GAVIE~\cite{liu2023mitigating} & 27.9 & 34.2 & \underline{39.5} \\ 
        Semantic Entropy~\cite{farquhar2024detecting} & \underline{43.6} & \underline{38.8} & 43.5 \\ 
        \rowcolor{gray!15} \textbf{Uncertainty-o (ours)} & \textbf{51.1} & \textbf{44.2} & \textbf{32.0} \\ 
        \bottomrule
    \end{tabular}
    % }
    \vspace{-5pt}
    
    \caption{
    \textbf{Generation Hallucination Detection Results.}
    We utilize StableDiffusion~\cite{rombach2022high}, VideoFusion~\cite{luo2023videofusion}, AnyGPT~\cite{zhan2024anygpt}, RGB2point~\cite{lee2024rgb2point} for image, video, audio, point cloud generation.
    % Uncertainty-o surpasses previous state-of-the-art methods in detecting hallucinations in various generative tasks, including image generation (Flickr), video generation (MSRVTT), audio generation (VCTK), and point cloud generation (Pix3D).
    }
    \label{tab:main_generation}
    \vspace{-10pt}
\end{table}

% We experiment with 8 different LMMs capable of understanding and generating various modalities of data. InternVL is a large vision-language model that takes image-text prompts and answers with text. OneLLM is proficient in processing various modal inputs, \eg, audio and point clouds, and outputs text-only responses. PointLLM is specifically designed for real-world point cloud understanding. VideoLLaMA takes videos as prompts and generates text answers. AnyGPT is capable of generating audio and visual data. HunyuanVideo is an industrial-level video generation model, taking text as a prompt. RGB2point realizes point cloud generation by taking a single image. StableDiffusion serves as a text-based image generation model.

\subsection{Multimodal Semantic Uncertainty} 

We quantify the variance of these sampled answers to capture the inherent uncertainty of the LMM. Instead of simply calculating variance based on format differences, we focus on variations in answer semantics.
To estimate the variance of sampled answer semantics, the most critical procedure lies in semantic checking between two generated answers. Given this challenge, especially for higher-level modalities, \eg, video, we propose utilizing an off-the-shelf LMM captioner to extract answer semantics and summarize the key points into text captions. Consequently, the challenging multimodal semantic checking is converted to the more manageable text semantic checking  (see Fig.~\ref{fig:method}). 

For those text captions, we directly prompt the LLM to iteratively check pairs of text answers. For texts with similar semantics, we cluster them into one group. After obtaining the semantic clusters of sampled answers, we analyze the discrete distribution of answer semantics and calculate the distribution entropy. Moreover, we normalize the calculated entropy compared to the maximum entropy decided by the sampling time, and obtain estimated LMM uncertainty, with higher values indicating greater LMM uncertainty:
\begin{equation}
u_{m} = - \sum_{i=1}^{n} p_i \log(p_i) \quad \text{where} \quad p_i = \frac{c_i}{C},
\end{equation}
where $c_i$ is answer count for $i$-th group, $C$ is the total answer count, $u_{m}$ is the estimated uncertainty for certain modal $m$.
Finally, uncertainties from all answer modalities is averaged to obtain final LMM uncertainty $u$.

\subsection{Uncertainty-Aware Downstream Task}

\noindent \textbf{Hallucination Detection.} We treat answers with low uncertainty as non-hallucinatory and those with high uncertainty as hallucinatory. In practice, we first infer LMM using the original prompt to obtain an initial answer. This answer is then compared with the ground truth to determine whether it contains hallucinations. Next, we apply the proposed multimodal prompt perturbation and multimodal semantic uncertainty to estimate LMM uncertainty for its initial answer. Finally, we evaluate the alignment between the estimated uncertainty and actual hallucination using 3 different metrics: Area under the Receiver Operating Characteristic Curve (AUROC)~\cite{farquhar2024detecting}, Area under the ‘Rejection Accuracy’ Curve (AURAC)~\cite{farquhar2024detecting}, and Expected Calibration Error (ECE)~\cite{xiong2023can}. AUROC and AURAC focus on ranking ability (higher is better), while ECE measures calibration (lower is better).

\vspace{1mm}
\noindent \textbf{Hallucination Mitigation.} We formulate hallucination mitigation as an uncertainty-aware, two-stage revision process. In the first stage, we prompt the LMM to generate an initial answer, which is then assigned an uncertainty score estimated by Uncertainty-o. After constructing an answer pool with corresponding uncertainties, we select the top-K answers with the highest uncertainty scores, as these are the most likely to contain hallucinations. In the second stage, we refine these high-uncertainty answers by prompting the LMM with the original context, initial answer, and uncertainty score. We explicitly tell the LMM that its initial answer has a high uncertainty score and instruct it to revise the response due to a potential hallucination. The revised answer is then used as the LMM final output:
\begin{equation}  
y_\text{final} = M(x', y_\text{initial}, u), \quad y_\text{initial} \in \mathbf{Y},
\end{equation}
where \( \mathbf{Y} \) contains selected answers with the highest uncertainty scores. \( x' \), \( y_\text{initial} \), and \( u \) refer to the updated prompt, initial answer, and estimated uncertainty, respectively. \( y_\text{final} \) is the revised answer. This iterative process helps the LMM refine its responses and improve accuracy.

\vspace{1mm}
\noindent \textbf{Uncertainty-Aware Chain-of-Thought.} We leverage Uncertainty-o to enhance the vanilla CoT process. Self-reflection is crucial for LMMs during CoT, and our estimated uncertainty serves as a guiding signal to trigger self-reflection. Specifically, at each reasoning step, in addition to generating responses, we also estimate answer uncertainty. In the subsequent step, we incorporate the previous answer and its uncertainty into the context, explicitly prompting LMM to reflect on its reasoning process when it finds uncertainty is high, as: 
\begin{equation}  
y_t, u_t = M(C_t), \quad C_{t+1} = C_t \cup \{y_t, u_t\} \,,
\end{equation}
where \( y_t \) and \( u_t \) is answer and uncertainty, given the context \( C_t \) at step \( t \). This uncertainty-aware procedure facilitates a more detailed and cautious reasoning process.

\section{Experiment}

%\subsection{Setting}

% \noindent \textbf{Benchmark.}
% We conduct experiments with 14 benchmarks. For the comprehension hallucination task, the benchmarks we adopt are typically QA-type. For images, we utilize MMVet, LLaVABench, and CoCoCap. For video benchmarks, we leverage MSRVTTQA, MSVDQA, and NextQA. For the audio modality, we use ClothV2 and AudioCaps. For point cloud benchmarks, we utilize PointCap and ModelNet. For generation hallucination detection, we use caption-type benchmarks and employ captions as generation prompts. For image generation, we utilize Flickr. For video generation, we utilize MSRVTT. For audio generation, we utilize VCTK. For point cloud generation, we utilize Pix3D, aiming for image-to-point cloud generation.

% \begin{table}[!t]

%     \centering
%     \fontsize{8}{9}\selectfont
%     \setlength{\tabcolsep}{1.5mm}
%     % \resizebox{0.8\linewidth}{!}{
%     \begin{tabular}{c|ccc}
%         \toprule
%         Method & AUROC$\uparrow$ & AURAC$\uparrow$ & ECE$\downarrow$ \\ 
%         \midrule
%         Cycle Consistency & 57.7 & 72.9 & 9.2 \\ 
%         \rowcolor{gray!15} Uncertainty-o & \textbf{59.5} & \textbf{74.5} & \textbf{8.3} \\ 
%         \bottomrule
%     \end{tabular}
%     % }
%     \vspace{-5pt}
    
%     \caption{
%     \textbf{Ablation of Cycle Consistency for Generation Hallucination Detection.} Uncertainty-o surpasses cycle consistency, as cycle consistency is constrained by the text similarity scoring capability of the LLM. We experiment on Flickr with Qwen2.5-7B.
%     }
%     \label{tab:ablation_cycle}
% \end{table}

\vspace{1mm}
% \noindent \textbf{LMM.}
% We experiment with 8 different LMMs capable of understanding and generating various modalities of data. InternVL is a large vision-language model that takes image-text prompts and answers with text. OneLLM is proficient in processing various modal inputs, \eg, audio and point clouds, and outputs text-only responses. PointLLM is specifically designed for real-world point cloud understanding. VideoLLaMA takes videos as prompts and generates text answers. AnyGPT is capable of generating audio and visual data. HunyuanVideo is an industrial-level video generation model, taking text as a prompt. RGB2point realizes point cloud generation by taking a single image. StableDiffusion serves as a text-based image generation model.

\vspace{1mm}
% \noindent \textbf{Implementation Details.}
% The hallucination detection procedure contains 3 key phases: (1) Initial answer obtaining. The tested LMM is first given the context to generate an initial output. We keep the variation hyper-parameter low at this stage, \eg, low temperature for LLM-based large models and high guidance scale for Diffusion models. The initial answer is compared with the ground truth to decide whether it contains hallucinations. (2) Uncertainty estimation. Initial contexts are perturbed by our proposed multimodal prompt perturbation, according to the modalities they contain. All modal prompts are perturbed simultaneously to various degrees. Different levels of perturbed prompts enable the sampling process. For sampled textual answers, an off-the-shelf LLM, Qwen2.5-7B, clusters them by semantics and calculates entropy as uncertainty. For other modalities' answers, \eg, images, we use OneLLM-7B as a captioner to convert those answers into concise captions. Then, the uncertainty calculation is simplified to vanilla textual semantic uncertainty. (3) Hallucination detection. With hallucination labels from (1) and estimated uncertainties from (2), we utilize AUROC \zznote{xxx}, AURAC\zznote{xxx}, and ECE\zznote{xxx} as hallucination detection metrics. These three metrics comprehensively evaluate the calibration degree between uncertainties and hallucinations.

\subsection{Comparison with State-of-the-Arts}

\noindent \textbf{Comprehension Hallucination Detection.} We compare Uncertainty-o with previous state-of-the-art methods in image, video, audio, and point cloud comprehension hallucination detection tasks (see Tab.~\ref{tab:main_comprehension}). Uncertainty-o yields significant and consistent performance improvements over strong baselines. This validates the general efficacy of the proposed multimodal prompt perturbation in better eliciting and capturing uncertainty, thereby facilitating reliable hallucination detection. Notably, Uncertainty-o surpasses baselines by 15.6\% (AUROC) on ClothoV2, 9.5\% (AURAC) on MMVet, 7.9\% (ECE) on ModelNet, and 4.1\% (AURAC) on NextQA. We also observe consistent improvement for closed-source LMMs, which validate the robustness of our Uncertainty-o (see Tab.~\ref{tab:main_close}). We further present a comparison of detection efficacy in the Medical Diagnosis and Embodied Robot domains (see Tab.~\ref{tab:main_safety}). We observe that Uncertainty-o effectively handles complexity in real-world applications and outperforms baselines by clear margins.
% Notably, Uncertainty-o surpasses baselines by 8.8\% in ECE on MIMICCXR and 10.6\% in AURAC on OpenEQA.
% These significant improvements demonstrate how properly designed prompt perturbation is crucial for refined uncertainty estimation across various modalities.

\noindent \textbf{Generation Hallucination Detection.} We also present results on generation hallucination detection tasks across image, video, audio, and point cloud modalities (see Tab.~\ref{tab:main_generation}). We adapt GAVIE~\cite{liu2023mitigating} here and extend text-only semantic entropy in~\cite{farquhar2024detecting} with our multimodal semantic uncertainty. We observe that Uncertainty-o consistently shows clear improvements over these baselines. For example, we Uncertainty-o achieves 23.2\% higher AUROC, 10\% higher AURAC in the point cloud generation task. These results validate that the interaction between multimodal prompt perturbation and multimodal semantic uncertainty can effectively achieve more accurate hallucination detection.

% Specifically, , while Uncertainty-o still outperforms this baseline. Moreover, GAVIE is limited by the scoring capability of off-the-shelf evaluators, and Uncertainty-o surpasses it by a clear margin.

\noindent \textbf{Hallucination Mitigation.} We report comprehension hallucination mitigation results on four video and audio benchmarks (see Tab.~\ref{tab:main_mitigation}). ``Direct inference'' accuracy is the result without our proposed uncertainty and is reproduced by ourselves. With our uncertainty-guided revision process, we observe consistent accuracy improvement. These results validate the efficacy of our estimated uncertainty in benefiting the hallucination mitigation process.

\begin{table}[!ht]
    \vspace{-20pt}

    \centering
    \fontsize{8}{9}\selectfont
    \setlength{\tabcolsep}{1.0mm}
    % \resizebox{0.8\linewidth}{!}{
    \begin{tabular}{l|cc}
        \toprule
        \multicolumn{3}{c}{\textit{Video Comprehension}} \\
        \midrule
        \multirow{2}{*}{VideoLLaMA~\cite{zhang2023video}} & MSRVTTQA~\cite{xu2017video} & MSVDQA~\cite{xu2017video} \\ 
        ~ & Acc. & Acc.\\ 
        \midrule
        Direct Inference & 62.9 & 70.5 \\
        \rowcolor{gray!15} \quad \textbf{+ Uncertainty-o} & \textbf{67.2} & \textbf{72.1} \\
        \midrule
        \multicolumn{3}{c}{\textit{Audio Comprehension}} \\
        \midrule
        \multirow{2}{*}{OneLLM~\cite{han2024onellm}} & ClothoAQA~\cite{lipping2022clotho} & ClothoV2~\cite{drossos2020clotho} \\ 
        ~ & Acc. & Acc. \\ 
        \midrule
        Direct Inference & 57.1 & 45.5 \\
        \rowcolor{gray!15} \quad \textbf{+ Uncertainty-o} & \textbf{59.2} & \textbf{51.9} \\
        \bottomrule
    \end{tabular}
    % }
    \vspace{-5pt}
    
    \caption{
    \textbf{Hallucination Mitigation Results.}
    Our uncertainty-guided revision effectively mitigates answer hallucination.
    }
    \label{tab:main_mitigation}
\end{table}

\begin{table}[!ht]
    \vspace{-6pt}
    \centering
    \fontsize{8}{9}\selectfont
    \setlength{\tabcolsep}{1.0mm}
    % \resizebox{0.8\linewidth}{!}{
    \begin{tabular}{l|cc}
        \toprule
        \multicolumn{3}{c}{\textit{Image Comprehension}} \\
        \midrule
        \multirow{2}{*}{InternVL~\cite{chen2024internvl}} & MMVet~\cite{yu2023mm} & LLaVABench~\cite{liu2023visual} \\ 
        ~ & \# Step & \# Step \\ 
        \midrule
        Vanilla CoT & 3.1 & 2.6 \\
        \rowcolor{gray!15} \textbf{Uncertainty-Aware CoT} & \textbf{5.4} & \textbf{4.1} \\
        \midrule
        \multicolumn{3}{c}{\textit{Point Cloud Comprehension}} \\
        \midrule
        \multirow{2}{*}{PointLLM~\cite{xu2024pointllm}} & ModelNet~\cite{wu20153d} & ShapeNet~\cite{chang2015shapenet} \\ 
        ~ & \# Step & \# Step \\ 
        \midrule
        Vanilla CoT & 2.7 & 3.5 \\
        \rowcolor{gray!15} \textbf{Uncertainty-Aware CoT} & \textbf{3.4} & \textbf{4.8} \\
        \bottomrule
    \end{tabular}
    % }
    \vspace{-5pt}
    
    \caption{
    \textbf{Uncertainty-Aware Chain-of-Thought Results.}
    Our estimated uncertainty enrich reasoning context, facilitating more thorough thinking process. We report average reasoning steps.
    }
    \label{tab:main_cot}
    \vspace{-10pt}
\end{table}

\noindent \textbf{Uncertainty-Aware CoT.} We show the comparison between Uncertainty-Aware CoT and vanilla CoT (see Tab.~\ref{tab:main_cot}). By incorporating uncertainty into the thinking context, LMMs generate a more comprehensive thinking process. On average, Uncertainty-Aware CoT leads to a noticeable increase in overall thinking length.

% \noindent \textbf{Safety-Critic Tasks.} 
% We further present a comparison of detection efficacy in the Medical Diagnosis and Embodied Robot domains (see Tab.~\ref{tab:main_safety}). We observe that Uncertainty-o effectively handles the complexity of these real-world applications and outperforms baselines by clear margins. Notably, Uncertainty-o surpasses baselines by 8.8\% in ECE on MIMICCXR and 10.6\% in AURAC on OpenEQA. As a result, the more reliable hallucination detection provided by Uncertainty-o can contribute to safer intelligent system deployments.

\begin{figure*}[!ht]
    \vspace{-15pt}
    \centering
    \includegraphics[width=\linewidth]{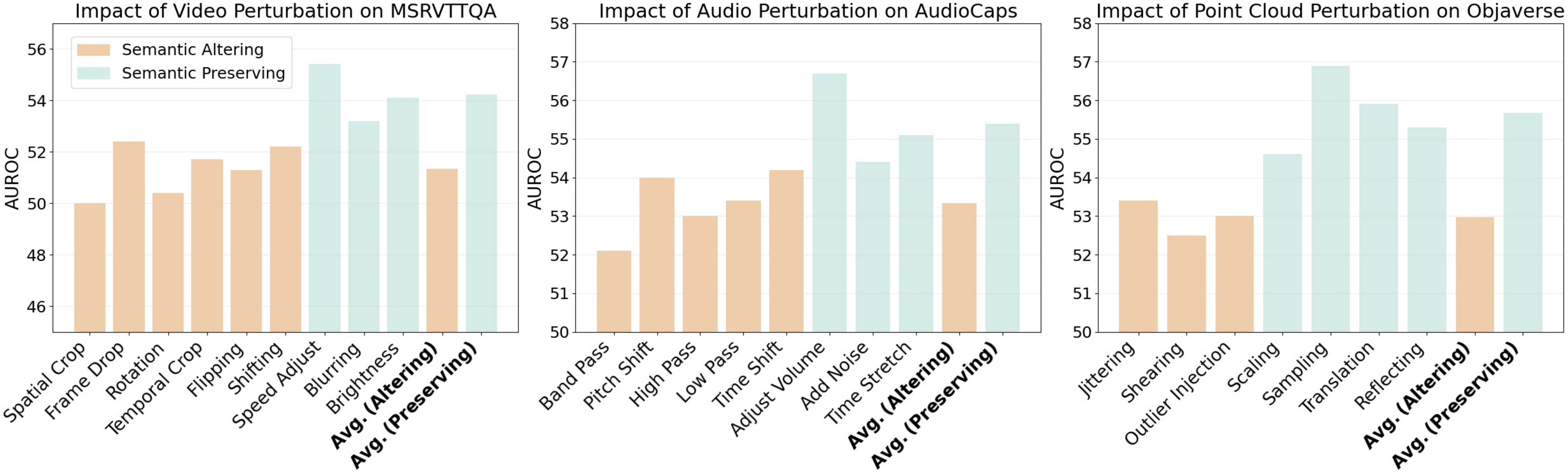}
    \vspace{-20pt}
    \caption{
    \textbf{Empirical Comparison of Different Prompt Perturbations.} On average, semantic-preserving perturbations are more effective for eliciting LMM uncertainty than semantic-altering ones. Results from comprehension hallucination detection for video, audio, point.
    }
    \label{fig:perturbation}
\end{figure*}

\subsection{Ablation Studies and Further Discussions}

\noindent \textbf{Empirical Study on Multimodal Prompt Perturbation.} We present an empirical comparison of 24 different prompt perturbations for video, audio, and point cloud (see Fig.~\ref{fig:perturbation}). For video perturbation, we find that perturbations that preserve the original video semantics, \eg, adjust video speed, achieve superior results. In contrast, perturbations that alter the original semantics, \eg, spatial cropping, lead to suboptimal results due to over-estimated uncertainty. For audio perturbation, a similar pattern is observed. We observe that perturbations that preserve the original audio semantics contribute more to reliable uncertainty estimation, \eg, adjusting volume. With the internal semantics intact and the external presentation changed, fluctuations in the sampled answers can directly reflect LMM uncertainty. For point cloud perturbation, a similar pattern also appears: operations that preserve the point cloud semantics are more beneficial for accurate hallucination detection.

\begin{table}[!ht]
    \vspace{-10pt}
    \centering
    \fontsize{8}{9}\selectfont
    \setlength{\tabcolsep}{1.5mm}
    % \resizebox{0.9\linewidth}{!}{
    \begin{tabular}{cccc|ccc}
        \toprule
        Text & Video & Audio & Point & AUROC$\uparrow$ & AURAC$\uparrow$ & ECE$\downarrow$ \\ 
        \midrule
        \ding{55} & \ding{55} & - & - & 51.6 & 35.1 & 14.7 \\ 
        \ding{51} & \ding{55} & - & - & 53.0 & 36.1 & 13.9 \\ 
        \ding{55} & \ding{51} & - & - & 54.1 & 37.5 & 12.9 \\ 
        \rowcolor{gray!15} \ding{51} & \ding{51} & - & - & \textbf{55.4} & \textbf{39.2} & \textbf{10.8} \\ 
        \midrule
        \midrule
        \ding{55} & - & \ding{55} & - & 50.9 & 42.4 & 16.1 \\ 
        \ding{51} & - & \ding{55} & - & 53.0 & 42.6 & 15.5 \\ 
        \ding{55} & - & \ding{51} & - & 52.6 & 44.9 & 12.9 \\ 
        \rowcolor{gray!15} \ding{51} & - & \ding{51} & - & \textbf{56.7} & \textbf{46.6} & \textbf{12.4} \\ 
        \midrule
        \midrule
        \ding{55} & - & - & \ding{55} & 50.1 & 38.5 & 26.0 \\ 
        \ding{51} & - & - & \ding{55} & 52.4 & 40.2 & 20.9 \\ 
        \ding{55} & - & - & \ding{51} & 54.1 & 41.6 & 19.1 \\ 
        \rowcolor{gray!15} \ding{51} & - & - & \ding{51} &  \textbf{56.9} & \textbf{43.8} & \textbf{18.7}  \\ 
        \bottomrule
    \end{tabular}
    % }
    \vspace{-5pt}
    
    \caption{
    \textbf{Ablation of Perturbation Combination.} We observe that combining perturbations for multimodal prompts yield better results. Single modal prompt perturbation could also yield performance improvements. Results from comprehension hallucination detection on MSRVTTQA~\cite{xu2017video}, AudioCaps~\cite{kim2019audiocaps}, and Objaverse~\cite{deitke2023objaverse}. Row with \colorbox{gray!15}{gray shading} is our default setting.
    }
    \label{tab:ablation_combination}
    \vspace{-10pt}
\end{table}

\begin{tcolorbox}[colback=gray!20, colframe=gray!80, sharp corners]
\textbf{Cookbook 1:} Try to use perturbation that largely preserves original prompt semantic.
\end{tcolorbox}

% \noindent \textbf{Video Perturbation.} We present ablation study of various video perturbations (see Tab.~\ref{tab:ablation_video_perturbation}). We find that perturbations that preserve the original video semantics achieve superior results on average. In contrast, perturbations that alter the original semantics lead to corresponding changes in the answers, resulting in over-estimated uncertainty. As a result, the semantic-preserving perturbations yield the optimal performance.

% \noindent \textbf{Audio Perturbation.} We report the ablation of different audio perturbations in Tab.~\ref{tab:ablation_audio_perturbation}. We observe that perturbations that preserve the original audio semantics contribute more to reliable hallucination detection. With the internal semantics intact and the external presentation changed, fluctuations in the sampled answers can directly reflect model uncertainty.

% \noindent \textbf{Point Cloud Perturbation.} We show the effectiveness of various point cloud perturbations on hallucination detection performance (see Tab.~\ref{tab:ablation_point_cloud_perturbation}). We observe a similar pattern to that seen in the video and audio domains: operations that preserve the point cloud semantics are more beneficial for accurate hallucination detection.

\noindent \textbf{Perturbation Combination.} We report the ablation of perturbation combinations in Tab.~\ref{tab:ablation_combination}. Concurrently perturbing multimodal prompts yields the best results. This approach effectively mines the uncertainty arising from each modality and contributes to capturing accurate LMM uncertainty.

\begin{tcolorbox}[colback=gray!20, colframe=gray!80, sharp corners]
\textbf{Cookbook 2:} Apply perturbation to prompt from different modalities simultaneously.
\end{tcolorbox}

\noindent \textbf{Pairing Order.} We present an ablation of pairing order in Tab.~\ref{tab:ablation_order}. For multimodal prompts perturbed to various degrees, pairing them with a similar degree in a progressive manner yields the best results. This approach creates a prompt set with challenge levels from low to high, effectively eliciting LMM uncertainty.

\begin{tcolorbox}[colback=gray!20, colframe=gray!80, sharp corners]
\textbf{Cookbook 3:} Perturb prompts from each modality to varying degrees and pair them in the progressive order.
\end{tcolorbox}

\begin{table}[!h]
    \vspace{-10pt}
    \centering
    \fontsize{8}{9}\selectfont
    \setlength{\tabcolsep}{2.4mm}
    % \resizebox{0.8\linewidth}{!}{
    \begin{tabular}{c|ccc}
        \toprule
        Pairing Order & AUROC$\uparrow$ & AURAC$\uparrow$ & ECE$\downarrow$ \\ 
        \midrule
        \rowcolor{gray!15} Progressive & \textbf{55.4} & \textbf{39.2} & \textbf{10.8} \\ 
        Random & 51.1 & 37.2 & 13.9 \\ 
        Shifted & 52.4 & 37.8 & 12.5 \\ 
        \bottomrule
    \end{tabular}
    % }
    \vspace{-5pt}
    
    \caption{
    \textbf{Pairing Order for Perturbed Prompts.}
    Pairing prompts with similar degrees of perturbation from different modalities yields the best results. Results from comprehension hallucination detection on MSRVTTQA. Our default setting is in \colorbox{gray!15}{gray}.
    }
    
    \label{tab:ablation_order}
\end{table}

\begin{table}[!h]
    \vspace{-20pt}

    \centering
    \fontsize{8}{9}\selectfont
    \setlength{\tabcolsep}{2mm}
    % \resizebox{0.8\linewidth}{!}{
    \begin{tabular}{c|ccc}
        \toprule
        Sampling Time & AUROC$\uparrow$ & AURAC$\uparrow$ & ECE$\downarrow$ \\ 
        \midrule
        2 & 50.6 & 35.1 & 15.9 \\ 
        3 & 54.2 & 36.9 & 12.4 \\ 
        \rowcolor{gray!15} 5 & \textbf{55.4} & \textbf{39.2} & \textbf{10.8} \\ 
        8 & 55.1 & 38.9 & 11.6 \\ 
        10 & 54.9 & 39.2 & 11.2 \\ 
        \bottomrule
    \end{tabular}
    % }
    \vspace{-5pt}
    
    \caption{
    \textbf{Ablation of Sampling Time.}
    Moderate sampling time enables the most effective uncertainty estimation. We report MSRVTTQA comprehension hallucination detection results. Our default setting is highlighted with \colorbox{gray!15}{gray shading}.
    }
    \label{tab:ablation_sampling_time}
    \vspace{0pt}
\end{table}

\noindent \textbf{Sampling Time.} We illustrate the ablation of sampling time during the uncertainty estimation process (see Tab.~\ref{tab:ablation_sampling_time}). With a moderate sampling time, \eg, 5 times, we achieve optimal results. A sampling time that is too small fails to effectively model LMM uncertainty, while excessively high sampling times inevitably lead to over-estimated uncertainty.

\begin{tcolorbox}[colback=gray!20, colframe=gray!80, sharp corners]
\textbf{Cookbook 4:} Try with a moderate perturbation time (sampling time), \eg, 5 times.
\end{tcolorbox}

% \noindent \textbf{Text Clustering Manner.} We report the ablation of text clustering methods (see Tab.~\ref{tab:ablation_text_clustering}). Clustering text based on its inherent semantics, rather than merely its lexical presentation, yields the best results.

% \noindent \textbf{Uncertainty - Error Rate Calibration.} We further analyze the calibration between estimated uncertainty and actual response error rate (see Fig.~\ref{fig:uncertainty_dis}). Uncertainty-o achieves the most calibrated uncertainty distribution with the answer error rate. In contrast, Confidence Elicitation suffers from the over-confidence problem, as its uncertainty is directly regressed by the LMM itself. Semantic Entropy achieves suboptimal calibration due to the lack of perturbation on multimodal prompts.

\begin{figure*}
    \centering
    \vspace{-15pt}
    \includegraphics[width=0.9\linewidth]{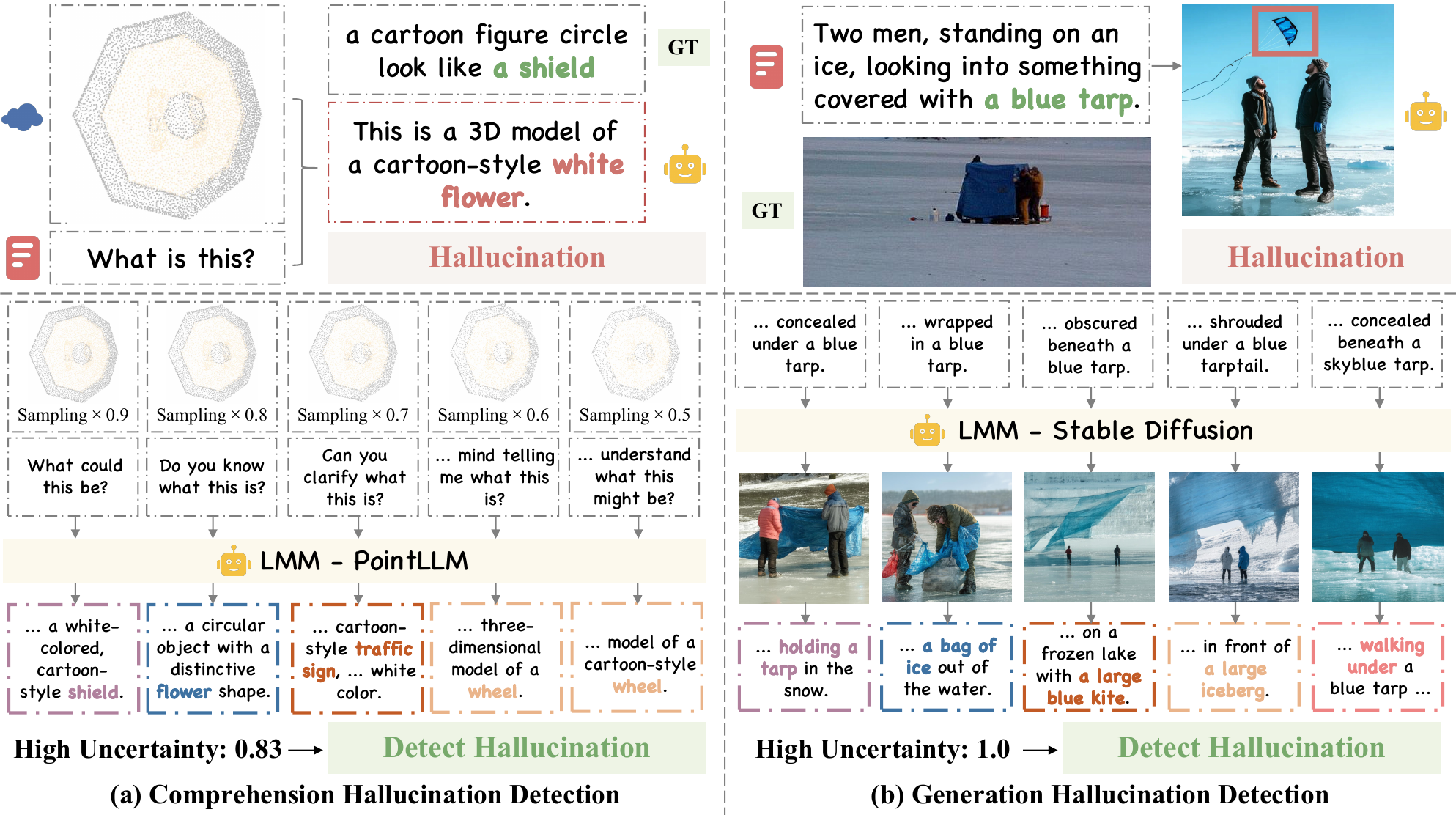}
    \vspace{-5pt}
    \caption{
    \textbf{Qualitative Results of Successful Hallucination Detection.} Uncertainty-o is proficient at accurately detecting multiple hallucinations, including comprehension (point cloud QA~\cite{deitke2023objaverse}) and generation tasks (image generation~\cite{plummer2015flickr30k}) based on the uncertainty value. %Cases are from Objaverse~\cite{deitke2023objaverse} and Flickr~\cite{plummer2015flickr30k}, respectively.
    }
    \label{fig:qualitative}
    \vspace{-10pt}
\end{figure*}

\noindent \textbf{Qualitative Results of Successful Hallucination Detection.} We present successful hallucination detection cases for both comprehension and generation scenarios in Fig.~\ref{fig:qualitative}. With the proposed multimodal prompt perturbation, Uncertainty-o achieves reliable uncertainty estimation, leading to accurate hallucination detection.

% \noindent \textbf{Successful Generation Hallucination Detection.} We also present case studies for generation hallucination detection in Fig.~\ref{fig:generation_case}. Our proposed multimodal semantic uncertainty effectively mines uncertainty from multimodal responses and contributes to the identification of generation hallucination.

\section{Related Work}

% My summary
\noindent \textbf{Large Multimodal Model.}
Large Multimodal Models (LMMs)~\cite{caffagni2024revolution} perceive and generate content across a wide range of modalities~\cite{caffagni2024revolution,wu2023multimodal,zhang2024approaching,zhang2024harnessing}, such as text~\cite{achiam2023gpt}, image~\cite{rombach2022high,fei2024vitron}, audio~\cite{zhan2024anygpt}, video~\cite{luo2023videofusion,fei2024video}, point cloud~\cite{xu2024pointllm}, and more. Fundamental efforts pioneer in building a universal-modal encoder by aligning various modalities to language~\cite{girdhar2023imagebind,lyu2024unibind}. On one hand, various works~\cite{liu2023visual,lin2023video,han2024onellm,xu2024pointllm} focus on enhancing multimodal comprehension ability~\cite{fei2024enhancing}, emphasizing the understanding of complexity~\cite{wu2024visual} and interaction~\cite{zhang2024prompt} between prompts from various modalities~\cite{wu2024can}. Another line of research focuses on generating multimodal content~\cite{rombach2022high,luo2023videofusion,zhan2024anygpt,lee2024rgb2point}, with an emphasis on the fidelity~\cite{song2024moma,li2024unimo} and diversity~\cite{fang2024puma,tevet2020evaluating} of multimodal responses. Moreover, recent works~\cite{wu2024next,zhan2024anygpt} try to build `Any-to-Any' capability~\cite{tang2023any,tang2024codi} in large models, where models can simultaneously comprehend complex contexts and generate high-fidelity content following those contexts.

% Hallucination of Multimodal Large Language Models: A Survey
\noindent \textbf{LMM Hallucination Detection.}
LMM Hallucination Detection~\cite{bai2024hallucination,huang2024visual} aims for detecting either context-misfollowing~\cite{huang2024survey} or factual mistakes~\cite{akhtar2023multimodal} in generated multimodal reponses. For external-evalutor-based approaches, powerful reponse scorers~\cite{liu2023mitigating,wang2023evaluation} or retrieved real-world fact~\cite{chen2024unified,ding2024retrieve} faciliate hallucination detection. For discrete rule-based checking, CHAIR~\cite{rohrbach2018object} employs the ratio of objects presented in the response relative to a ground truth object list to identify hallucinatory answers, which is limited to fixed object classes. Building on CHAIR, POPE~\cite{li2023evaluating} enhances the prompting technique by focusing on Yes-or-No questions, simplifying process and improving stability. Distinct from existing works, Uncertainty-o harnesses LMM uncertainty as intrinsic indicator of hallucination responses.

% https://github.com/Ruiyang-061X/Awesome-MLLM-Uncertainty
\noindent \textbf{Uncertainty in LMM.}
MAP~\cite{ji2022map} proposes a probabilistic distribution encoder to model uncertainty during large model pretraining. VL-Uncertainty~\cite{zhang2024vl} leverages semantic-equivalent perturbations of visual and textual prompts to better capture MLLM uncertainty. DropoutDecoding~\cite{fang2024uncertainty} treats token distribution discrepancies from the average token distribution as uncertainty, dropping tokens with high uncertainty. Calibration-MLLM~\cite{chen2024unveiling} utilizes calibration between different stages of MLLM, such as before and after visual fine-tuning, to reveal uncertainty. IDK~\cite{cohen2025don} introduces an additional special `\texttt{I Dont Know}' token, and quantify uncertainty based on predicted probability of this token. UAL~\cite{wang2024uncertainty} utilizes uncertainty for large model alignment and improves token convergence in feature space.

\section{Conclusion}

% Recently, with the widespread and rapid deployment of Large Multimodal Models (LMMs) in real-world applications, concerns about the safety of these models have emerged. Among the various indicators of LMM reliability, we argue that their intrinsic uncertainty remains under-explored. In this paper, we attempt to answer three critical questions: (1) How can we systematically unveil the uncertainty of LMMs capable of perceiving and generating multimodal content? (2) How can we model the correspondence between complexity in multimodal prompts and heightened LMM uncertainty? (3) How can we extract uncertainty from multimodal responses, beyond just text answers? To address these questions, (1) we propose Uncertainty-o, a unified and model-agnostic framework to unveil uncertainty in LMMs, capable of perceiving and generating 5 modalities. (2) We introduce multimodal prompt perturbation and empirically demonstrate that semantic-preserving perturbations are more effective for capturing LMM uncertainty. (3) We introduce multimodal semantic uncertainty, effectively mining uncertainty from multimodal generated responses. Extensive experiments and analysis validate superiority of Uncertainty-o in accurately estimating LMM uncertainty and facilitating various downstream tasks.

In this paper, we study LMMs reliability and explore the intrinsic uncertainty in LMMs. We propose Uncertainty-o, a model-agnostic framework that unveils uncertainty in LMMs and simultaneously processes five different modalities (visual, auditory, textual, video, and point clouds). Specifically, we first introduce multimodal prompt perturbation and empirically demonstrate that semantic-preserving perturbations are more effective for capturing LMM uncertainty. Then, we introduce multimodal semantic uncertainty based on the imposed perturbations, effectively mining uncertainty from multimodal generated responses. Extensive experiments and analysis validate superiority of Uncertainty-o in accurately estimating LMM uncertainty and facilitating various downstream tasks.

{
    \small
    \bibliographystyle{ieeenat_fullname}
    \bibliography{main}
}

% WARNING: do not forget to delete the supplementary pages from your submission 
\input{sec/X_suppl}

\end{document}

%% file: sec/X_suppl.tex
\clearpage
\setcounter{page}{1}
\maketitlesupplementary

\section{Dicussion}

\noindent \textbf{Why is semantic-equivalent perturbation more effective than inequivalent perturbation?}
In semantic-equivalent perturbation, the semantics of all perturbed prompts remain the same. As a result, any fluctuation in answers of LMM directly reflects its uncertainty. On the other hand, semantic-inequivalent perturbation alters the original semantics, and changes in the answers are inevitable, making it difficult to accurately reflect the inherent uncertainty of the LMM.

\vspace{1mm}

\noindent \textbf{Why is prompt perturbation superior to vanilla sampling for capturing LMM uncertainty?}
Multimodal prompt perturbation challenges the LMM with prompts that vary at different levels, with larger variations of answers indicating higher uncertainty. In contrast, vanilla sampling typically uses a high temperature during sampling, which tends to induce high uncertainty across all samples. For instance, the LMM could be highly confident in its answer, but the high temperature inevitably causes answer variation, leading to an artificially high uncertainty. 

\vspace{1mm}

\noindent \textbf{Can multimodal semantic uncertainty extend to more modalities?}
Yes. Uncertainty-o proposes a general pipeline for LMM uncertainty estimation. This pipeline is orthogonal to and can benefit from iterations of LMM. With a more general LMM, which is capable of parsing the semantic content of additional modalities into textual descriptions, Uncertainty-o can seamlessly estimate uncertainty for those answers in more modalities.  

\vspace{1mm}

\noindent \textbf{Why not directly utilize LMM to check whether two generated multimodal responses are semantically equivalent?}
Current LMMs still lack proficient ability for multimodal semantic checking. We have attempted two methods to utilize LMM for direct multimodal data semantic checking: (1) Combine two multimodal answers into one and prompt the LMM to check whether the answers are similar. For example, merge two point clouds into one and conduct shifting so that they do not overlap. However, current LMMs show unsatisfactory performance. (2) Directly input two answers into LMM and prompt it to check. We find that many LMMs still do not support multi-input, such as understanding two point clouds simultaneously.

\section{Setting}

\noindent \textbf{Benchmark.}
We conduct experiments with 18 benchmarks. For the comprehension hallucination task, the benchmarks we adopt are typically QA-type. For images, we utilize MMVet, LLaVABench, and CoCoCap. For video benchmarks, we leverage MSRVTTQA, MSVDQA, and NextQA. For the audio modality, we use ClothV2 and AudioCaps. For point cloud benchmarks, we utilize PointCap and ModelNet. For generation hallucination detection, we use caption-type benchmarks and employ captions as generation prompts. For image generation, we utilize Flickr. For video generation, we utilize MSRVTT. For audio generation, we utilize VCTK. For point cloud generation, we utilize Pix3D, aiming for image-to-point cloud generation. For the safety-critical task, we utilize MIMICCXR and OpenEQA. MIMICCXR is a benchmark of chest X-ray images annotated with medical diagnoses by radiologists. OpenEQA is an in-house video benchmark from an embodied robot's perspective, accompanied by text-based questions.

\vspace{1mm}

\noindent \textbf{LMM.}
We experiment with 10 different LMMs capable of understanding and generating various modalities of data. For closed-source LMMs, we utilize GPT-4o and QwenVLMax. For open-source LMMs, InternVL is a large vision-language model that takes image-text prompts and answers with text. VideoLLaMA takes videos as prompts and generates text answers. OneLLM is proficient in processing various modal inputs, \eg, audio and point clouds, and outputs text-only responses. PointLLM is specifically designed for real-world point cloud understanding. StableDiffusion serves as a text-based image generation model. VideoFusion is an efficient video generation model, taking text as a prompt. AnyGPT is capable of generating audio and visual data. RGB2point realizes point cloud generation by taking a single image.

\vspace{1mm}

\noindent \textbf{Implementation Details.}
The hallucination detection procedure contains 3 key phases: (1) Initial answer obtaining. The tested LMM is first given the context to generate an initial output. We keep the variation hyper-parameter low at this stage, \eg, low temperature for LLM-based large models and high guidance scale for Diffusion models. The initial answer is compared with the ground truth to decide whether it contains hallucinations. (2) Uncertainty estimation. Initial contexts are perturbed by our proposed multimodal prompt perturbation, according to the modalities they contain. All modal prompts are perturbed simultaneously to various degrees. Different levels of perturbed prompts enable the sampling process. For sampled textual answers, an off-the-shelf LLM, Qwen2.5-7B, clusters them by semantics and calculates entropy as uncertainty. For other modalities' answers, \eg, images, we use OneLLM-7B as a captioner to convert those answers into concise captions. Then, the uncertainty calculation is simplified to vanilla textual semantic uncertainty. (3) Hallucination detection. With hallucination labels from (1) and estimated uncertainties from (2), we utilize AUROC, AURAC, and ECE as hallucination detection metrics. These three metrics comprehensively evaluate the calibration degree between uncertainties and hallucinations. For hallucination mitigation, we follow a two-stage procedure: (1) We first prompt the LMM with initial context to obtain an initial answer and calculate its uncertainty using Uncertainty-o. (2) We then select the top 50\% of answers with the highest uncertainty, as these are most likely to contain hallucinations. For each of these answers, we utilize the following prompt: `Prompt: \texttt{\$X}, Initial Answer: \texttt{\$Y}, Your answer has a high uncertainty score of \texttt{\$U}, which ranges from 0 to 1. Could you improve your answer and revise it to be more accurate? \texttt{\$X}, \texttt{\$Y}, and \texttt{\$U} refer to the actual content of original prompt, inital answer and estimated uncertainty. Finally, we use the revised answer as the final output from the LMM. For uncertainty-aware CoT, we treat it as a multi-round thinking process. In the first round, we append the original prompt with `Let's think step-by-step. Now, provide your first step of the answer:' We then estimate the LMM uncertainty with Uncertainty-o. From the second step onward, in addition to the original prompt, we append all previous answers and their corresponding uncertainty scores. At the end of the prompt, we add `Respond with `Finish.' when you think you have solved the question.' We check answer at every step for the presence of `Finish.' and exit the loop once we find it.

\section{More Ablations}

\begin{table}[!t]
    \centering
    \fontsize{8}{9}\selectfont
    \setlength{\tabcolsep}{2.4mm}
    % \resizebox{0.8\linewidth}{!}{
    \begin{tabular}{c|ccc}
        \toprule
        Text Clustering & AUROC$\uparrow$ & AURAC$\uparrow$ & ECE$\downarrow$ \\ 
        \midrule
        Lexical & 48.0 & 33.2 & 12.7 \\ 
        \rowcolor{gray!15} Semantic & \textbf{55.4} & \textbf{39.2} & \textbf{10.8} \\ 
        \bottomrule
    \end{tabular}
    % }
    \caption{
    \textbf{Ablation of Text Clustering.} During semantic uncertainty calculation, clustering based on semantics is more effective than clustering based merely on lexical presentation.
    }
    \label{tab:ablation_text_clustering}
\end{table}

\begin{table}[!t]

    \centering
    \fontsize{8}{9}\selectfont
    \setlength{\tabcolsep}{2.4mm}
    % \resizebox{0.8\linewidth}{!}{
    \begin{tabular}{c|ccc}
        \toprule
        Method & AUROC$\uparrow$ & AURAC$\uparrow$ & ECE$\downarrow$ \\ 
        \midrule
        Cycle Consistency & 57.7 & 72.9 & 9.2 \\ 
        \rowcolor{gray!15} Uncertainty-o & \textbf{59.5} & \textbf{74.5} & \textbf{8.3} \\ 
        \bottomrule
    \end{tabular}
    % }
    
    \caption{
    \textbf{Ablation of Cycle Consistency for Generation Hallucination Detection.} Uncertainty-o surpasses cycle consistency, as cycle consistency is constrained by the text similarity scoring capability of the LLM. We experiment on Flickr with StableDiffusion.
    }
    \label{tab:ablation_cycle}
    \vspace{-10pt}
\end{table}

\noindent \textbf{Text Clustering Manner.}
We report the ablation of text clustering methods (see Tab.~\ref{tab:ablation_text_clustering}). Clustering text based on its inherent semantics, rather than merely its lexical presentation, yields the best results.

\noindent \textbf{Cycle Consistency.}
We also experiment with cycle consistency for detecting hallucinations in image generation settings (see Tab.~\ref{tab:ablation_cycle}). We observe that Uncertainty-o surpasses this baseline method, which is limited by the scoring ability of the LLM for text pair similarity.

\vspace{1mm}

% \section{Perturbation Visualization}

% \begin{figure*}
%     \centering
%     \includegraphics[width=\linewidth]{perturbation_point_cloud.pdf}
%     \caption{
%     \textbf{Visualization of Point Cloud Perturbation.}
%     }
%     \label{fig:perturbation_point_cloud}
% \end{figure*}

\section{Detailed Proof for Proportional Theorem}
\label{sec:proof}

\subsection{Definitions and Assumptions}

\begin{enumerate}
    \item \textbf{Sampling Process:} The large model $M$ give predictions concurrently for perturbed prompts $x_{i}$ and $x_{j}$. For those input prompts, the predictions from the large model is $y_{\text{i}} = M(x_{\text{i}})$, and $y_{\text{j}} = M(x_{\text{j}})$, respectively.
    \item \textbf{Uncertainty:} We focus on the large model's \textit{Epistemic Uncertainty}, which is the uncertainty in the large model parameters. Assume the large model parameters $\theta$ are random variables with a prior distribution $P(\theta)$.
    \item \textbf{Prediction Difference:} Define the prediction difference $D(x)$ as:
    \[
    D(x) = \| y_{\text{i}} - y_{\text{j}} \|,
    \]
    where $\|\cdot\|$ denotes semantic space distance.
\end{enumerate}

\subsection{Mathematical Derivation}

\noindent \textbf{Large Model’s Predictive Distribution.} Assume the large model’s output is a probability distribution $P(y|x, \theta)$, where $y$ is the response, $x$ is the prompt, and $\theta$ are the model parameters.

\noindent \textbf{Posterior Predictive Distribution.} According to Bayes’ theorem, the posterior predictive distribution of the large model can be expressed as:
\[
P(y|x) = \int P(y|x, \theta) P(\theta|x) d\theta,
\]
where $P(\theta|x)$ is the posterior distribution of the large model parameters.

\noindent \textbf{Epistemic Uncertainty.} The uncertainty in the parameters can be measured by the variance of the posterior distribution:
\[
\text{Var}(\theta|x) = \mathbb{E}_{\theta|x}[(\theta - \mathbb{E}_{\theta|x}[\theta])^2] = \mathbb{E}_{\theta|x}[\theta^2] - (\mathbb{E}_{\theta|x}[\theta])^2.
\]

\noindent \textbf{Prediction Difference and Epistemic Uncertainty.} To relate the prediction difference $D(x)$ to parameter uncertainty, we need to consider the predictions from the perturbed prompts. Assume the parameters during $i$-th and $j$-th prompt perturbation are $\theta_{\text{i}}$ and $\theta_{\text{j}}$, respectively, and they have the same prior distribution, i.e., $P(\theta_{\text{i}}) = P(\theta_{\text{j}})$.

\noindent \textbf{Predictions from Perturbed Prompts.} The predictions from perturbed prompts can be expressed as:
\[
y_{\text{i}} = \mathbb{E}_{\theta_{\text{i}}|x} \left[ P(y|x, \theta_{\text{i}}) \right],
\]
\[
y_{\text{j}} = \mathbb{E}_{\theta_{\text{j}}|x} \left[ P(y|x, \theta_{\text{j}}) \right].
\]

\noindent \textbf{Expression for Prediction Difference.} Assume the difference in predictions can be approximated by a First-order Taylor Expansion:
\[
y_{\text{i}} - y_{\text{j}} \approx \mathbb{E}_{\theta|x} \left[ \nabla_{\theta} P(y|x, \theta) \cdot (\theta_{\text{i}} - \theta_{\text{j}}) \right],
\]
where $\nabla_{\theta} P(y|x, \theta)$ is the gradient of $P(y|x, \theta)$ with respect to $\theta$.

Thus, the prediction difference $D(x)$ can be expressed as:
\[
D(x) = \| y_{\text{i}} - y_{\text{j}} \| \approx \mathbb{E}_{\theta|x} \left[ \| \nabla_{\theta} P(y|x, \theta) \cdot (\theta_{\text{i}} - \theta_{\text{j}}) \| \right].
\]

\noindent \textbf{Relationship Between Prediction Difference from Prompt Perturbation and Epistemic Uncertainty.} To simplify the analysis, assume $\theta_{\text{i}}$ and $\theta_{\text{j}}$ are independently and identically distributed (i.i.d.). Then:
\[
\mathbb{E}_{\theta|x} \left[ \| \nabla_{\theta} P(y|x, \theta) \cdot (\theta_{\text{i}} - \theta_{\text{j}}) \|^2 \right] \approx 
\]
\[
\mathbb{E}_{\theta|x} \left[ \| \nabla_{\theta} P(y|x, \theta) \|^2 \right] \cdot \mathbb{E}_{\theta|x} \left[ (\theta_{\text{i}} - \theta_{\text{j}})^2 \right].
\]
With $\mathbb{E}_{\theta|x} \left[ (\theta_{\text{i}} - \theta_{\text{j}})^2 \right] = 2\left(\mathbb{E}_{\theta|x} [\theta^2] - (\mathbb{E}_{\theta|x} [\theta])^2\right) = \\ 2 \cdot \text{Var}(\theta|x)
$, we have:
\[
\mathbb{E}_{\theta|x} \left[ \| \nabla_{\theta} P(y|x, \theta) \cdot (\theta_{\text{i}} - \theta_{\text{j}}) \|^2 \right] \approx 
\]
\[
2 \cdot \mathbb{E}_{\theta|x} \left[ \| \nabla_{\theta} P(y|x, \theta) \|^2 \right] \cdot \text{Var}(\theta|x).
\]

Assuming $k = \mathbb{E}_{\theta|x} \left[ \| \nabla_{\theta} P(y|x, \theta) \|^2 \right]$, which is positive, we get:
\[
\mathbb{E}_{\theta|x} \left[ \| \nabla_{\theta} P(y|x, \theta) \cdot (\theta_{\text{i}} - \theta_{\text{j}}) \|^2 \right] \approx 2k \cdot \text{Var}(\theta|x).
\]

Thus, the prediction difference $D(x)$ from prompt perturbation can be expressed as:
\[
D(x) \approx \sqrt{2k} \cdot \sqrt{\text{Var}(\theta|x)}.
\]

Simplifying further, we obtain:
\[
D(x) \propto \sqrt{\text{Var}(\theta|x)}.
\]

\subsection{Final Theorem}

From the above derivation, we have shown that the prediction difference $D(x)$ from prompt perturbation is proportional to the square root of the model's epistemic uncertainty $\sqrt{\text{Var}(\theta|x)}$. Therefore, the prediction difference $D(x)$ can serve as a measure of the model’s epistemic uncertainty for a given sample.

%% file: main.bbl
\begin{thebibliography}{83}
\providecommand{\natexlab}[1]{#1}
\providecommand{\url}[1]{\texttt{#1}}
\expandafter\ifx\csname urlstyle\endcsname\relax
  \providecommand{\doi}[1]{doi: #1}\else
  \providecommand{\doi}{doi: \begingroup \urlstyle{rm}\Url}\fi

\bibitem[Achiam et~al.(2023)Achiam, Adler, Agarwal, Ahmad, Akkaya, Aleman, Almeida, Altenschmidt, Altman, Anadkat, et~al.]{achiam2023gpt}
Josh Achiam, Steven Adler, Sandhini Agarwal, Lama Ahmad, Ilge Akkaya, Florencia~Leoni Aleman, Diogo Almeida, Janko Altenschmidt, Sam Altman, Shyamal Anadkat, et~al.
\newblock Gpt-4 technical report.
\newblock \emph{arXiv:2303.08774}, 2023.

\bibitem[Aichberger et~al.(2024)Aichberger, Schweighofer, Ielanskyi, and Hochreiter]{aichberger2024many}
Lukas Aichberger, Kajetan Schweighofer, Mykyta Ielanskyi, and Sepp Hochreiter.
\newblock How many opinions does your llm have? improving uncertainty estimation in nlg.
\newblock In \emph{ICLR 2024 Workshop on Secure and Trustworthy Large Language Models}, 2024.

\bibitem[Akhtar et~al.(2023)Akhtar, Schlichtkrull, Guo, Cocarascu, Simperl, and Vlachos]{akhtar2023multimodal}
Mubashara Akhtar, Michael Schlichtkrull, Zhijiang Guo, Oana Cocarascu, Elena Simperl, and Andreas Vlachos.
\newblock Multimodal automated fact-checking: A survey.
\newblock \emph{arXiv:2305.13507}, 2023.

\bibitem[Amayuelas et~al.(2023)Amayuelas, Wong, Pan, Chen, and Wang]{amayuelas2023knowledge}
Alfonso Amayuelas, Kyle Wong, Liangming Pan, Wenhu Chen, and William Wang.
\newblock Knowledge of knowledge: Exploring known-unknowns uncertainty with large language models.
\newblock \emph{arXiv:2305.13712}, 2023.

\bibitem[Auletta et~al.(2023)Auletta, Kallen, di~Bernardo, and Richardson]{auletta2023predicting}
Fabrizia Auletta, Rachel~W Kallen, Mario di Bernardo, and Michael~J Richardson.
\newblock Predicting and understanding human action decisions during skillful joint-action using supervised machine learning and explainable-ai.
\newblock \emph{Scientific Reports}, 13\penalty0 (1):\penalty0 4992, 2023.

\bibitem[Bai et~al.(2023)Bai, Bai, Yang, Wang, Tan, Wang, Lin, Zhou, and Zhou]{bai2023qwenvlversatilevisionlanguagemodel}
Jinze Bai, Shuai Bai, Shusheng Yang, Shijie Wang, Sinan Tan, Peng Wang, Junyang Lin, Chang Zhou, and Jingren Zhou.
\newblock Qwen-vl: A versatile vision-language model for understanding, localization, text reading, and beyond, 2023.

\bibitem[Bai et~al.(2024)Bai, Wang, Xiao, He, Han, Zhang, and Shou]{bai2024hallucination}
Zechen Bai, Pichao Wang, Tianjun Xiao, Tong He, Zongbo Han, Zheng Zhang, and Mike~Zheng Shou.
\newblock Hallucination of multimodal large language models: A survey.
\newblock \emph{arXiv:2404.18930}, 2024.

\bibitem[Brown et~al.(2020)Brown, Mann, Ryder, Subbiah, Kaplan, Dhariwal, Neelakantan, Shyam, Sastry, Askell, et~al.]{brown2020language}
Tom Brown, Benjamin Mann, Nick Ryder, Melanie Subbiah, Jared~D Kaplan, Prafulla Dhariwal, Arvind Neelakantan, Pranav Shyam, Girish Sastry, Amanda Askell, et~al.
\newblock Language models are few-shot learners.
\newblock \emph{Advances in neural information processing systems}, 33:\penalty0 1877--1901, 2020.

\bibitem[Caffagni et~al.(2024)Caffagni, Cocchi, Barsellotti, Moratelli, Sarto, Baraldi, Cornia, and Cucchiara]{caffagni2024revolution}
Davide Caffagni, Federico Cocchi, Luca Barsellotti, Nicholas Moratelli, Sara Sarto, Lorenzo Baraldi, Marcella Cornia, and Rita Cucchiara.
\newblock The revolution of multimodal large language models: a survey.
\newblock \emph{arXiv:2402.12451}, 2024.

\bibitem[Callaway et~al.(2022)Callaway, van Opheusden, Gul, Das, Krueger, Griffiths, and Lieder]{callaway2022rational}
Frederick Callaway, Bas van Opheusden, Sayan Gul, Priyam Das, Paul~M Krueger, Thomas~L Griffiths, and Falk Lieder.
\newblock Rational use of cognitive resources in human planning.
\newblock \emph{Nature Human Behaviour}, 6\penalty0 (8):\penalty0 1112--1125, 2022.

\bibitem[Carolan et~al.(2024)Carolan, Fennelly, and Smeaton]{carolan2024review}
Kilian Carolan, Laura Fennelly, and Alan~F Smeaton.
\newblock A review of multi-modal large language and vision models.
\newblock \emph{arXiv:2404.01322}, 2024.

\bibitem[Chang et~al.(2015)Chang, Funkhouser, Guibas, Hanrahan, Huang, Li, Savarese, Savva, Song, Su, et~al.]{chang2015shapenet}
Angel~X Chang, Thomas Funkhouser, Leonidas Guibas, Pat Hanrahan, Qixing Huang, Zimo Li, Silvio Savarese, Manolis Savva, Shuran Song, Hao Su, et~al.
\newblock Shapenet: An information-rich 3d model repository.
\newblock \emph{arXiv:1512.03012}, 2015.

\bibitem[Chen et~al.(2015)Chen, Fang, Lin, Vedantam, Gupta, Doll{\'a}r, and Zitnick]{chen2015microsoft}
Xinlei Chen, Hao Fang, Tsung-Yi Lin, Ramakrishna Vedantam, Saurabh Gupta, Piotr Doll{\'a}r, and C~Lawrence Zitnick.
\newblock Microsoft coco captions: Data collection and evaluation server.
\newblock \emph{arXiv preprint arXiv:1504.00325}, 2015.

\bibitem[Chen et~al.(2024{\natexlab{a}})Chen, Wang, Xue, Zhang, Yang, Li, Shen, Liang, Gu, and Chen]{chen2024unified}
Xiang Chen, Chenxi Wang, Yida Xue, Ningyu Zhang, Xiaoyan Yang, Qiang Li, Yue Shen, Lei Liang, Jinjie Gu, and Huajun Chen.
\newblock Unified hallucination detection for multimodal large language models.
\newblock \emph{arXiv:2402.03190}, 2024{\natexlab{a}}.

\bibitem[Chen et~al.(2024{\natexlab{b}})Chen, Hu, He, Deng, Zhang, and Hong]{chen2024unveiling}
Zijun Chen, Wenbo Hu, Guande He, Zhijie Deng, Zheng Zhang, and Richang Hong.
\newblock Unveiling uncertainty: A deep dive into calibration and performance of multimodal large language models.
\newblock \emph{arXiv:2412.14660}, 2024{\natexlab{b}}.

\bibitem[Chen et~al.(2024{\natexlab{c}})Chen, Wu, Wang, Su, Chen, Xing, Zhong, Zhang, Zhu, Lu, et~al.]{chen2024internvl}
Zhe Chen, Jiannan Wu, Wenhai Wang, Weijie Su, Guo Chen, Sen Xing, Muyan Zhong, Qinglong Zhang, Xizhou Zhu, Lewei Lu, et~al.
\newblock Internvl: Scaling up vision foundation models and aligning for generic visual-linguistic tasks.
\newblock In \emph{Proceedings of the IEEE/CVF conference on computer vision and pattern recognition}, pages 24185--24198, 2024{\natexlab{c}}.

\bibitem[Cohen et~al.(2024)Cohen, Dobler, Biran, and de~Melo]{cohen2025don}
Roi Cohen, Konstantin Dobler, Eden Biran, and Gerard de Melo.
\newblock I don't know: Explicit modeling of uncertainty with an [idk] token.
\newblock \emph{Advances in Neural Information Processing Systems}, 37:\penalty0 10935--10958, 2024.

\bibitem[Collins et~al.(2023)Collins, Barker, Espinosa~Zarlenga, Raman, Bhatt, Jamnik, Sucholutsky, Weller, and Dvijotham]{collins2023human}
Katherine~Maeve Collins, Matthew Barker, Mateo Espinosa~Zarlenga, Naveen Raman, Umang Bhatt, Mateja Jamnik, Ilia Sucholutsky, Adrian Weller, and Krishnamurthy Dvijotham.
\newblock Human uncertainty in concept-based ai systems.
\newblock In \emph{Proceedings of the 2023 AAAI/ACM Conference on AI, Ethics, and Society}, pages 869--889, 2023.

\bibitem[Deitke et~al.(2023)Deitke, Schwenk, Salvador, Weihs, Michel, VanderBilt, Schmidt, Ehsani, Kembhavi, and Farhadi]{deitke2023objaverse}
Matt Deitke, Dustin Schwenk, Jordi Salvador, Luca Weihs, Oscar Michel, Eli VanderBilt, Ludwig Schmidt, Kiana Ehsani, Aniruddha Kembhavi, and Ali Farhadi.
\newblock Objaverse: A universe of annotated 3d objects.
\newblock In \emph{CVPR}, pages 13142--13153, 2023.

\bibitem[Ding et~al.(2024)Ding, Pang, Wei, Shen, and Cheng]{ding2024retrieve}
Hanxing Ding, Liang Pang, Zihao Wei, Huawei Shen, and Xueqi Cheng.
\newblock Retrieve only when it needs: Adaptive retrieval augmentation for hallucination mitigation in large language models.
\newblock \emph{arXiv:2402.10612}, 2024.

\bibitem[Drossos et~al.(2020)Drossos, Lipping, and Virtanen]{drossos2020clotho}
Konstantinos Drossos, Samuel Lipping, and Tuomas Virtanen.
\newblock Clotho: An audio captioning dataset.
\newblock In \emph{ICASSP 2020-2020 IEEE International Conference on Acoustics, Speech and Signal Processing (ICASSP)}, pages 736--740. IEEE, 2020.

\bibitem[Fang et~al.(2024{\natexlab{a}})Fang, Duan, Wang, Li, Tian, Zeng, Zhao, Dai, Li, and Liu]{fang2024puma}
Rongyao Fang, Chengqi Duan, Kun Wang, Hao Li, Hao Tian, Xingyu Zeng, Rui Zhao, Jifeng Dai, Hongsheng Li, and Xihui Liu.
\newblock Puma: Empowering unified mllm with multi-granular visual generation.
\newblock \emph{arXiv:2410.13861}, 2024{\natexlab{a}}.

\bibitem[Fang et~al.(2024{\natexlab{b}})Fang, Yang, Chen, Zhao, and Zhou]{fang2024uncertainty}
Yixiong Fang, Ziran Yang, Zhaorun Chen, Zhuokai Zhao, and Jiawei Zhou.
\newblock From uncertainty to trust: Enhancing reliability in vision-language models with uncertainty-guided dropout decoding.
\newblock \emph{arXiv:2412.06474}, 2024{\natexlab{b}}.

\bibitem[Farquhar et~al.(2024)Farquhar, Kossen, Kuhn, and Gal]{farquhar2024detecting}
Sebastian Farquhar, Jannik Kossen, Lorenz Kuhn, and Yarin Gal.
\newblock Detecting hallucinations in large language models using semantic entropy.
\newblock \emph{Nature}, 630\penalty0 (8017):\penalty0 625--630, 2024.

\bibitem[Fei et~al.(2024{\natexlab{a}})Fei, Wu, Ji, Zhang, Zhang, Lee, and Hsu]{fei2024video}
Hao Fei, Shengqiong Wu, Wei Ji, Hanwang Zhang, Meishan Zhang, Mong-Li Lee, and Wynne Hsu.
\newblock Video-of-thought: Step-by-step video reasoning from perception to cognition.
\newblock \emph{arXiv preprint arXiv:2501.03230}, 2024{\natexlab{a}}.

\bibitem[Fei et~al.(2024{\natexlab{b}})Fei, Wu, Zhang, Chua, and Yan]{fei2024vitron}
Hao Fei, Shengqiong Wu, Hanwang Zhang, Tat-Seng Chua, and Shuicheng Yan.
\newblock Vitron: A unified pixel-level vision llm for understanding, generating, segmenting, editing.
\newblock \emph{arXiv preprint arXiv:2412.19806}, 2024{\natexlab{b}}.

\bibitem[Fei et~al.(2024{\natexlab{c}})Fei, Wu, Zhang, Zhang, Chua, and Yan]{fei2024enhancing}
Hao Fei, Shengqiong Wu, Meishan Zhang, Min Zhang, Tat-Seng Chua, and Shuicheng Yan.
\newblock Enhancing video-language representations with structural spatio-temporal alignment.
\newblock \emph{IEEE Transactions on Pattern Analysis and Machine Intelligence}, 2024{\natexlab{c}}.

\bibitem[Girdhar et~al.(2023)Girdhar, El-Nouby, Liu, Singh, Alwala, Joulin, and Misra]{girdhar2023imagebind}
Rohit Girdhar, Alaaeldin El-Nouby, Zhuang Liu, Mannat Singh, Kalyan~Vasudev Alwala, Armand Joulin, and Ishan Misra.
\newblock Imagebind: One embedding space to bind them all.
\newblock In \emph{CVPR}, pages 15180--15190, 2023.

\bibitem[Griffin et~al.(2023)Griffin, Kleinberg, Mozes, Mai, Vau, Caldwell, and Mavor-Parker]{griffin2023large}
Lewis Griffin, Bennett Kleinberg, Maximilian Mozes, Kimberly Mai, Maria Do~Mar Vau, Matthew Caldwell, and Augustine Mavor-Parker.
\newblock Large language models respond to influence like humans.
\newblock In \emph{Proceedings of the First Workshop on Social Influence in Conversations (SICon 2023)}, pages 15--24, 2023.

\bibitem[Han et~al.(2024)Han, Gong, Zhang, Wang, Zhang, Lin, Qiao, Gao, and Yue]{han2024onellm}
Jiaming Han, Kaixiong Gong, Yiyuan Zhang, Jiaqi Wang, Kaipeng Zhang, Dahua Lin, Yu Qiao, Peng Gao, and Xiangyu Yue.
\newblock Onellm: One framework to align all modalities with language.
\newblock In \emph{CVPR}, pages 26584--26595, 2024.

\bibitem[Ho et~al.(2020)Ho, Jain, and Abbeel]{ho2020denoising}
Jonathan Ho, Ajay Jain, and Pieter Abbeel.
\newblock Denoising diffusion probabilistic models.
\newblock \emph{Advances in neural information processing systems}, 33:\penalty0 6840--6851, 2020.

\bibitem[Huang and Zhang(2024)]{huang2024survey}
Jiaxing Huang and Jingyi Zhang.
\newblock A survey on evaluation of multimodal large language models.
\newblock \emph{arXiv:2408.15769}, 2024.

\bibitem[Huang et~al.(2024)Huang, Liu, Guo, and Gong]{huang2024visual}
Wen Huang, Hongbin Liu, Minxin Guo, and Neil~Zhenqiang Gong.
\newblock Visual hallucinations of multi-modal large language models.
\newblock \emph{arXiv:2402.14683}, 2024.

\bibitem[Hurst et~al.(2024)Hurst, Lerer, Goucher, Perelman, Ramesh, Clark, Ostrow, Welihinda, Hayes, Radford, et~al.]{hurst2024gpt}
Aaron Hurst, Adam Lerer, Adam~P Goucher, Adam Perelman, Aditya Ramesh, Aidan Clark, AJ Ostrow, Akila Welihinda, Alan Hayes, Alec Radford, et~al.
\newblock Gpt-4o system card.
\newblock \emph{arXiv preprint arXiv:2410.21276}, 2024.

\bibitem[Ji et~al.(2022)Ji, Wang, Gong, Zhang, Zhu, Wang, Zhang, Sakai, and Yang]{ji2022map}
Yatai Ji, Junjie Wang, Yuan Gong, Lin Zhang, Yanru Zhu, Hongfa Wang, Jiaxing Zhang, Tetsuya Sakai, and Yujiu Yang.
\newblock Map: Multimodal uncertainty-aware vision-language pre-training model.
\newblock \emph{arXiv:2210.05335}, 2022.

\bibitem[Johnson et~al.(2019)Johnson, Pollard, Berkowitz, Greenbaum, Lungren, Deng, Mark, and Horng]{johnson2019mimic}
Alistair~EW Johnson, Tom~J Pollard, Seth~J Berkowitz, Nathaniel~R Greenbaum, Matthew~P Lungren, Chih-ying Deng, Roger~G Mark, and Steven Horng.
\newblock Mimic-cxr, a de-identified publicly available database of chest radiographs with free-text reports.
\newblock \emph{Scientific data}, 6\penalty0 (1):\penalty0 317, 2019.

\bibitem[Ke et~al.(2024)Ke, Tong, Cheng, and Peng]{ke2024exploring}
Luoma Ke, Song Tong, Peng Cheng, and Kaiping Peng.
\newblock Exploring the frontiers of llms in psychological applications: A comprehensive review.
\newblock \emph{arXiv:2401.01519}, 2024.

\bibitem[Kim et~al.(2019)Kim, Kim, Lee, and Kim]{kim2019audiocaps}
Chris~Dongjoo Kim, Byeongchang Kim, Hyunmin Lee, and Gunhee Kim.
\newblock Audiocaps: Generating captions for audios in the wild.
\newblock In \emph{Proceedings of the 2019 Conference of the North American Chapter of the Association for Computational Linguistics: Human Language Technologies, Volume 1 (Long and Short Papers)}, pages 119--132, 2019.

\bibitem[Kreuk et~al.(2022)Kreuk, Synnaeve, Polyak, Singer, D{\'e}fossez, Copet, Parikh, Taigman, and Adi]{kreuk2022audiogen}
Felix Kreuk, Gabriel Synnaeve, Adam Polyak, Uriel Singer, Alexandre D{\'e}fossez, Jade Copet, Devi Parikh, Yaniv Taigman, and Yossi Adi.
\newblock Audiogen: Textually guided audio generation.
\newblock \emph{arXiv:2209.15352}, 2022.

\bibitem[Lee and Benes(2024)]{lee2024rgb2point}
Jae~Joong Lee and Bedrich Benes.
\newblock Rgb2point: 3d point cloud generation from single rgb images.
\newblock \emph{arXiv:2407.14979}, 2024.

\bibitem[Li et~al.(2024)Li, Xu, Liu, and Xiao]{li2024unimo}
Wei Li, Xue Xu, Jiachen Liu, and Xinyan Xiao.
\newblock Unimo-g: Unified image generation through multimodal conditional diffusion.
\newblock \emph{arXiv:2401.13388}, 2024.

\bibitem[Li et~al.(2023)Li, Du, Zhou, Wang, Zhao, and Wen]{li2023evaluating}
Yifan Li, Yifan Du, Kun Zhou, Jinpeng Wang, Wayne~Xin Zhao, and Ji-Rong Wen.
\newblock Evaluating object hallucination in large vision-language models.
\newblock \emph{arXiv:2305.10355}, 2023.

\bibitem[Lin et~al.(2023)Lin, Ye, Zhu, Cui, Ning, Jin, and Yuan]{lin2023video}
Bin Lin, Yang Ye, Bin Zhu, Jiaxi Cui, Munan Ning, Peng Jin, and Li Yuan.
\newblock Video-llava: Learning united visual representation by alignment before projection.
\newblock \emph{arXiv:2311.10122}, 2023.

\bibitem[Lipping et~al.(2022)Lipping, Sudarsanam, Drossos, and Virtanen]{lipping2022clotho}
Samuel Lipping, Parthasaarathy Sudarsanam, Konstantinos Drossos, and Tuomas Virtanen.
\newblock Clotho-aqa: A crowdsourced dataset for audio question answering.
\newblock In \emph{2022 30th European Signal Processing Conference (EUSIPCO)}, pages 1140--1144. IEEE, 2022.

\bibitem[Liu et~al.(2023{\natexlab{a}})Liu, Lin, Li, Wang, Yacoob, and Wang]{liu2023mitigating}
Fuxiao Liu, Kevin Lin, Linjie Li, Jianfeng Wang, Yaser Yacoob, and Lijuan Wang.
\newblock Mitigating hallucination in large multi-modal models via robust instruction tuning.
\newblock \emph{arXiv:2306.14565}, 2023{\natexlab{a}}.

\bibitem[Liu et~al.(2023{\natexlab{b}})Liu, Li, Wu, and Lee]{liu2023visual}
Haotian Liu, Chunyuan Li, Qingyang Wu, and Yong~Jae Lee.
\newblock Visual instruction tuning.
\newblock \emph{NeurIPS}, 36:\penalty0 34892--34916, 2023{\natexlab{b}}.

\bibitem[Luo et~al.(2023)Luo, Chen, Zhang, Huang, Wang, Shen, Zhao, Zhou, and Tan]{luo2023videofusion}
Zhengxiong Luo, Dayou Chen, Yingya Zhang, Yan Huang, Liang Wang, Yujun Shen, Deli Zhao, Jingren Zhou, and Tieniu Tan.
\newblock Videofusion: Decomposed diffusion models for high-quality video generation.
\newblock \emph{arXiv:2303.08320}, 2023.

\bibitem[Lyu et~al.(2024)Lyu, Zheng, Zhou, and Wang]{lyu2024unibind}
Yuanhuiyi Lyu, Xu Zheng, Jiazhou Zhou, and Lin Wang.
\newblock Unibind: Llm-augmented unified and balanced representation space to bind them all.
\newblock In \emph{CVPR}, pages 26752--26762, 2024.

\bibitem[Majumdar et~al.(2024)Majumdar, Ajay, Zhang, Putta, Yenamandra, Henaff, Silwal, Mcvay, Maksymets, Arnaud, et~al.]{majumdar2024openeqa}
Arjun Majumdar, Anurag Ajay, Xiaohan Zhang, Pranav Putta, Sriram Yenamandra, Mikael Henaff, Sneha Silwal, Paul Mcvay, Oleksandr Maksymets, Sergio Arnaud, et~al.
\newblock Openeqa: Embodied question answering in the era of foundation models.
\newblock In \emph{CVPR}, pages 16488--16498, 2024.

\bibitem[Nichol et~al.(2022)Nichol, Jun, Dhariwal, Mishkin, and Chen]{nichol2022point}
Alex Nichol, Heewoo Jun, Prafulla Dhariwal, Pamela Mishkin, and Mark Chen.
\newblock Point-e: A system for generating 3d point clouds from complex prompts.
\newblock \emph{arXiv:2212.08751}, 2022.

\bibitem[Peterson et~al.(2019)Peterson, Battleday, Griffiths, and Russakovsky]{peterson2019human}
Joshua~C Peterson, Ruairidh~M Battleday, Thomas~L Griffiths, and Olga Russakovsky.
\newblock Human uncertainty makes classification more robust.
\newblock In \emph{ICCV}, pages 9617--9626, 2019.

\bibitem[Pi et~al.(2024)Pi, Han, Zhang, Xie, Pan, Lian, Dong, Zhang, and Zhang]{pi2024mllm}
Renjie Pi, Tianyang Han, Jianshu Zhang, Yueqi Xie, Rui Pan, Qing Lian, Hanze Dong, Jipeng Zhang, and Tong Zhang.
\newblock Mllm-protector: Ensuring mllm's safety without hurting performance.
\newblock \emph{arXiv:2401.02906}, 2024.

\bibitem[Plummer et~al.(2015)Plummer, Wang, Cervantes, Caicedo, Hockenmaier, and Lazebnik]{plummer2015flickr30k}
Bryan~A Plummer, Liwei Wang, Chris~M Cervantes, Juan~C Caicedo, Julia Hockenmaier, and Svetlana Lazebnik.
\newblock Flickr30k entities: Collecting region-to-phrase correspondences for richer image-to-sentence models.
\newblock In \emph{ICCV}, pages 2641--2649, 2015.

\bibitem[Radford et~al.(2023)Radford, Kim, Xu, Brockman, McLeavey, and Sutskever]{radford2023robust}
Alec Radford, Jong~Wook Kim, Tao Xu, Greg Brockman, Christine McLeavey, and Ilya Sutskever.
\newblock Robust speech recognition via large-scale weak supervision.
\newblock In \emph{ICML}, pages 28492--28518. PMLR, 2023.

\bibitem[Rohrbach et~al.(2018)Rohrbach, Hendricks, Burns, Darrell, and Saenko]{rohrbach2018object}
Anna Rohrbach, Lisa~Anne Hendricks, Kaylee Burns, Trevor Darrell, and Kate Saenko.
\newblock Object hallucination in image captioning.
\newblock \emph{arXiv:1809.02156}, 2018.

\bibitem[Rombach et~al.(2022)Rombach, Blattmann, Lorenz, Esser, and Ommer]{rombach2022high}
Robin Rombach, Andreas Blattmann, Dominik Lorenz, Patrick Esser, and Bj{\"o}rn Ommer.
\newblock High-resolution image synthesis with latent diffusion models.
\newblock In \emph{CVPR}, pages 10684--10695, 2022.

\bibitem[Shorinwa et~al.(2024)Shorinwa, Mei, Lidard, Ren, and Majumdar]{shorinwa2024survey}
Ola Shorinwa, Zhiting Mei, Justin Lidard, Allen~Z Ren, and Anirudha Majumdar.
\newblock A survey on uncertainty quantification of large language models: Taxonomy, open research challenges, and future directions.
\newblock \emph{arXiv:2412.05563}, 2024.

\bibitem[Song et~al.(2024)Song, Zhu, Liu, Yan, Elgammal, and Yang]{song2024moma}
Kunpeng Song, Yizhe Zhu, Bingchen Liu, Qing Yan, Ahmed Elgammal, and Xiao Yang.
\newblock Moma: Multimodal llm adapter for fast personalized image generation.
\newblock In \emph{ECCV}, pages 117--132, 2024.

\bibitem[Sun et~al.(2018)Sun, Wu, Zhang, Zhang, Zhang, Xue, Tenenbaum, and Freeman]{sun2018pix3d}
Xingyuan Sun, Jiajun Wu, Xiuming Zhang, Zhoutong Zhang, Chengkai Zhang, Tianfan Xue, Joshua~B Tenenbaum, and William~T Freeman.
\newblock Pix3d: Dataset and methods for single-image 3d shape modeling.
\newblock In \emph{CVPR}, pages 2974--2983, 2018.

\bibitem[Tang et~al.(2023)Tang, Yang, Zhu, Zeng, and Bansal]{tang2023any}
Zineng Tang, Ziyi Yang, Chenguang Zhu, Michael Zeng, and Mohit Bansal.
\newblock Any-to-any generation via composable diffusion.
\newblock \emph{Advances in Neural Information Processing Systems}, 36:\penalty0 16083--16099, 2023.

\bibitem[Tang et~al.(2024)Tang, Yang, Khademi, Liu, Zhu, and Bansal]{tang2024codi}
Zineng Tang, Ziyi Yang, Mahmoud Khademi, Yang Liu, Chenguang Zhu, and Mohit Bansal.
\newblock Codi-2: In-context interleaved and interactive any-to-any generation.
\newblock In \emph{CVPR}, pages 27425--27434, 2024.

\bibitem[Tevet and Berant(2020)]{tevet2020evaluating}
Guy Tevet and Jonathan Berant.
\newblock Evaluating the evaluation of diversity in natural language generation.
\newblock \emph{arXiv:2004.02990}, 2020.

\bibitem[Wang et~al.(2023)Wang, Zhou, Xu, Shi, Zhao, Xu, Ye, Yan, Zhang, Zhu, et~al.]{wang2023evaluation}
Junyang Wang, Yiyang Zhou, Guohai Xu, Pengcheng Shi, Chenlin Zhao, Haiyang Xu, Qinghao Ye, Ming Yan, Ji Zhang, Jihua Zhu, et~al.
\newblock Evaluation and analysis of hallucination in large vision-language models.
\newblock \emph{arXiv:2308.15126}, 2023.

\bibitem[Wang et~al.(2024)Wang, Zheng, Ding, Zhang, Lin, and Tao]{wang2024uncertainty}
Yikun Wang, Rui Zheng, Liang Ding, Qi Zhang, Dahua Lin, and Dacheng Tao.
\newblock Uncertainty aware learning for language model alignment.
\newblock \emph{arXiv:2406.04854}, 2024.

\bibitem[Wu et~al.(2023)Wu, Gan, Chen, Wan, and Philip]{wu2023multimodal}
Jiayang Wu, Wensheng Gan, Zefeng Chen, Shicheng Wan, and S~Yu Philip.
\newblock Multimodal large language models: A survey.
\newblock In \emph{2023 IEEE International Conference on Big Data (BigData)}, pages 2247--2256. IEEE, 2023.

\bibitem[Wu et~al.(2024{\natexlab{a}})Wu, Zhang, Xia, Li, Xia, Chang, Yu, Kim, Rossi, Zhang, et~al.]{wu2024visual}
Junda Wu, Zhehao Zhang, Yu Xia, Xintong Li, Zhaoyang Xia, Aaron Chang, Tong Yu, Sungchul Kim, Ryan~A Rossi, Ruiyi Zhang, et~al.
\newblock Visual prompting in multimodal large language models: A survey.
\newblock \emph{arXiv:2409.15310}, 2024{\natexlab{a}}.

\bibitem[Wu et~al.(2024{\natexlab{b}})Wu, Fei, Qu, Ji, and Chua]{wu2024next}
Shengqiong Wu, Hao Fei, Leigang Qu, Wei Ji, and Tat-Seng Chua.
\newblock Next-gpt: Any-to-any multimodal llm.
\newblock In \emph{ICML}, 2024{\natexlab{b}}.

\bibitem[Wu et~al.(2024{\natexlab{c}})Wu, Lian, Gonzalez, Li, and Darrell]{wu2024self}
Tsung-Han Wu, Long Lian, Joseph~E Gonzalez, Boyi Li, and Trevor Darrell.
\newblock Self-correcting llm-controlled diffusion models.
\newblock In \emph{CVPR}, pages 6327--6336, 2024{\natexlab{c}}.

\bibitem[Wu et~al.(2024{\natexlab{d}})Wu, Shen, Shan, Song, Wang, Zhang, Feng, Cheng, Chen, Xiong, et~al.]{wu2024can}
Xixi Wu, Yifei Shen, Caihua Shan, Kaitao Song, Siwei Wang, Bohang Zhang, Jiarui Feng, Hong Cheng, Wei Chen, Yun Xiong, et~al.
\newblock Can graph learning improve planning in llm-based agents?
\newblock In \emph{The Thirty-eighth Annual Conference on Neural Information Processing Systems}, 2024{\natexlab{d}}.

\bibitem[Wu et~al.(2015)Wu, Song, Khosla, Yu, Zhang, Tang, and Xiao]{wu20153d}
Zhirong Wu, Shuran Song, Aditya Khosla, Fisher Yu, Linguang Zhang, Xiaoou Tang, and Jianxiong Xiao.
\newblock 3d shapenets: A deep representation for volumetric shapes.
\newblock In \emph{CVPR}, pages 1912--1920, 2015.

\bibitem[Xiao et~al.(2021)Xiao, Shang, Yao, and Chua]{xiao2021next}
Junbin Xiao, Xindi Shang, Angela Yao, and Tat-Seng Chua.
\newblock Next-qa: Next phase of question-answering to explaining temporal actions.
\newblock In \emph{CVPR}, pages 9777--9786, 2021.

\bibitem[Xiong et~al.(2023)Xiong, Hu, Lu, Li, Fu, He, and Hooi]{xiong2023can}
Miao Xiong, Zhiyuan Hu, Xinyang Lu, Yifei Li, Jie Fu, Junxian He, and Bryan Hooi.
\newblock Can llms express their uncertainty? an empirical evaluation of confidence elicitation in llms.
\newblock \emph{arXiv:2306.13063}, 2023.

\bibitem[Xu et~al.(2017)Xu, Zhao, Xiao, Wu, Zhang, He, and Zhuang]{xu2017video}
Dejing Xu, Zhou Zhao, Jun Xiao, Fei Wu, Hanwang Zhang, Xiangnan He, and Yueting Zhuang.
\newblock Video question answering via gradually refined attention over appearance and motion.
\newblock In \emph{Proceedings of the 25th ACM international conference on Multimedia}, pages 1645--1653, 2017.

\bibitem[Xu et~al.(2016)Xu, Mei, Yao, and Rui]{xu2016msr}
Jun Xu, Tao Mei, Ting Yao, and Yong Rui.
\newblock Msr-vtt: A large video description dataset for bridging video and language.
\newblock In \emph{CVPR}, pages 5288--5296, 2016.

\bibitem[Xu et~al.(2024)Xu, Wang, Wang, Chen, Pang, and Lin]{xu2024pointllm}
Runsen Xu, Xiaolong Wang, Tai Wang, Yilun Chen, Jiangmiao Pang, and Dahua Lin.
\newblock Pointllm: Empowering large language models to understand point clouds.
\newblock In \emph{ECCV}, pages 131--147, 2024.

\bibitem[Yamagishi et~al.(2019)Yamagishi, Veaux, and MacDonald]{Yamagishi2019CSTRVC}
Junichi Yamagishi, Christophe Veaux, and Kirsten MacDonald.
\newblock Cstr vctk corpus: English multi-speaker corpus for cstr voice cloning toolkit (version 0.92).
\newblock 2019.

\bibitem[Yu et~al.(2023)Yu, Yang, Li, Wang, Lin, Liu, Wang, and Wang]{yu2023mm}
Weihao Yu, Zhengyuan Yang, Linjie Li, Jianfeng Wang, Kevin Lin, Zicheng Liu, Xinchao Wang, and Lijuan Wang.
\newblock Mm-vet: Evaluating large multimodal models for integrated capabilities.
\newblock \emph{arXiv:2308.02490}, 2023.

\bibitem[Zhan et~al.(2024)Zhan, Dai, Ye, Zhou, Zhang, Liu, Zhang, Yuan, Zhang, Li, et~al.]{zhan2024anygpt}
Jun Zhan, Junqi Dai, Jiasheng Ye, Yunhua Zhou, Dong Zhang, Zhigeng Liu, Xin Zhang, Ruibin Yuan, Ge Zhang, Linyang Li, et~al.
\newblock Anygpt: Unified multimodal llm with discrete sequence modeling.
\newblock \emph{arXiv:2402.12226}, 2024.

\bibitem[Zhang et~al.(2023)Zhang, Li, and Bing]{zhang2023video}
Hang Zhang, Xin Li, and Lidong Bing.
\newblock Video-llama: An instruction-tuned audio-visual language model for video understanding.
\newblock \emph{arXiv preprint arXiv:2306.02858}, 2023.

\bibitem[Zhang et~al.(2024{\natexlab{a}})Zhang, Zhang, Yu, and Zheng]{zhang2024approaching}
Ruiyang Zhang, Hu Zhang, Hang Yu, and Zhedong Zheng.
\newblock Approaching outside: scaling unsupervised 3d object detection from 2d scene.
\newblock In \emph{ECCV}, pages 249--266, 2024{\natexlab{a}}.

\bibitem[Zhang et~al.(2024{\natexlab{b}})Zhang, Zhang, Yu, and Zheng]{zhang2024harnessing}
Ruiyang Zhang, Hu Zhang, Hang Yu, and Zhedong Zheng.
\newblock Harnessing uncertainty-aware bounding boxes for unsupervised 3d object detection.
\newblock \emph{arXiv:2408.00619}, 2024{\natexlab{b}}.

\bibitem[Zhang et~al.(2024{\natexlab{c}})Zhang, Zhang, and Zheng]{zhang2024vl}
Ruiyang Zhang, Hu Zhang, and Zhedong Zheng.
\newblock Vl-uncertainty: Detecting hallucination in large vision-language model via uncertainty estimation.
\newblock \emph{arXiv:2411.11919}, 2024{\natexlab{c}}.

\bibitem[Zhang et~al.(2024{\natexlab{d}})Zhang, Qian, Peng, Liu, and Jia]{zhang2024prompt}
Yuechen Zhang, Shengju Qian, Bohao Peng, Shu Liu, and Jiaya Jia.
\newblock Prompt highlighter: Interactive control for multi-modal llms.
\newblock In \emph{CVPR}, pages 13215--13224, 2024{\natexlab{d}}.

\end{thebibliography}
